Τμήμα Μηχανικών Η/Υ & Πληροφορικής

Μεταπτυχιακή Διπλωματική Εργασία

## «Ανάπτυξη αποδοτικού αλγορίθμου συνεργατικής διήθησης με χρήση των διανυσμάτων Lanczos για το πρόβλημα των top-N συστάσεων σε συστήματα μεγάλου όγκου δεδομένων»

*της*
Μαρίας Καλαντζή

*Επιβλέπων:*
Καθ. Γιάννης Γαροφαλάκης

14 Ιουνίου 2016

# Πανεπιστήμιο Πατρών

## Τμήμα Μηχανικών Η/Υ & Πληροφορικής

Μεταπτυχιακή Διπλωματική Εργασία

«Ανάπτυξη αποδοτικού αλγορίθμου συνεργατικής διήθησης με χρήση των διανυσμάτων Lanczos για το πρόβλημα των top-N συστάσεων σε συστήματα μεγάλου όγκου δεδομένων»

Μαρία Καλαντζή, Α.Μ. 824

| Καθηγητής | Καθηγητής | Αναπληρωτής Καθηγητής |
|---|---|---|
| Γιάννης Γαροφαλάκης | Αθανάσιος Τσακαλίδης | Ιωάννης Χατζηλυγερούδης |
| ………………………… | ………………………… | ………………………… |
| ………………………… | ………………………… | ………………………… |
| ………………………… | ………………………… | ………………………… |

14 Ιουνίου 2016

*Στο Θάνο ...*

# *Περίληψη*

Σκοπός της παρούσας διπλωματικής εργασίας είναι η μελέτη και ανάπτυξη ενός νέου αλγοριθμικού πλαισίου *Συνεργατικής Διήθησης* (**C**ollaborative **F**iltering) για την παραγωγή συστάσεων, και ειδικότερα για το πρόβλημα των *top-N* συστάσεων.

Προτείνουμε λοιπόν τον *Lanczos Latent Factor Recommender* (LLFR), έναν νέο CF αλγόριθμο, φιλικό στη διαχείριση μεγάλου όγκου δεδομένων για το πρόβλημα των top-N συστάσεων. Χρησιμοποιώντας μία προσέγγιση υπολογιστικά αποδοτική η οποία βασίζεται στη μέθοδο Lanczos, ο LLFR μειώνει τη διάσταση του προβλήματος κατασκευάζοντας ένα latent factor μοντέλο, το οποίο μπορεί άμεσα να αξιοποιηθεί για την παραγωγή προσωποποιημένων διανυσμάτων κατάταξης στο χώρο αντικειμένων.

Μια σειρά πειραμάτων σε πραγματικά σύνολα δεδομένων (`MovieLens10M, Yahoo!Music`) διαφορετικών επιπέδων πυκνότητας υποδεικνύουν ότι ο LLFR αποδίδει καλύτερα συγκριτικά με άλλες γνωστές μεθόδους top-N συστάσεων, τόσο από υπολογιστική όσο και ποιοτική σκοπιά. Επιπλέον, τα πειραματικά μας αποτελέσματα δείχνουν ότι αυτό το προβάδισμα στην απόδοση, συγκρινόμενο με ανταγωνιστικές μεθόδους, αυξάνεται καθώς τα δεδομένα γίνονται αραιότερα, δηλαδή στις περιπτώσεις όπου δεν υπάρχουν αρκετά διαθέσιμα δεδομένα προκειμένου να αναγνωριστούν ομοιότητες και εντέλει να παραχθούν αξιόπιστες συστάσεις.

Πιο συγκεκριμένα, ο LLFR αποδίδει καλύτερα τόσο στην περίπτωση όπου η αραιότητα είναι γενικευμένη – όπως στο *New Community Problem*, το οποίο συναντάται στα πραγματικά συστήματα κατά τα αρχικά τους στάδια όπου δεν υπάρχουν αρκετά δεδομένα ακόμα στο σύστημα – όσο και στην ιδιαίτερα ενδιαφέρουσα περίπτωση όπου η αραιότητα εντοπίζεται τοπικά σε ένα μικρό κομμάτι των δεδομένων – όπως στο *New Users Problem*, το οποίο συναντάται κατά την εισαγωγή νέων χρηστών σε ένα υπάρχον σύστημα, όπου ακριβώς επειδή αυτοί οι χρήστες είναι νέοι δεν έχουν προλάβει να βαθμολογήσουν αντικείμενα.

**Λέξεις Κλειδιά**

Συστήματα Συστάσεων, Συνεργατική Διήθηση, Αραιότητα, Top-N, Recommender Systems, Collaborative Filtering, Sparsity, Lanczos Method, Dimensionality Reduction

# *Abstract*

The purpose if this master's thesis is to study and develop a new algorithmic framework for ***C**ollaboartive **F**iltering* to produce recommendations in the *top-N* recommendation problem.

Thus, we propose *Lanczos Latent Factor Recommender* (LLFR); a novel "big data friendly" collaborative filtering algorithm for top-N recommendation. Using a computationally efficient Lanczos-based procedure, LLFR builds a low dimensional item similarity model, that can be readily exploited to produce personalized ranking vectors over the item space.

A number of experiments on real datasets (`MovieLens10M, Yahoo!Music`) at different density levels indicate that LLFR outperforms other state-of-the-art top-N recommendation methods from a computational as well as a qualitative perspective. Our experimental results also show that its relative performance gains, compared to competing methods, increase as the data get sparser, where there is not enough data for the system to uncover similarities and generate reliable recommendations.

More specifically, this is true both when the sparsity is generalized – as in the *New Community Problem*, a very common problem faced by real recommender systems in their beginning stages, when there is not sufficient number of ratings for the collaborative filtering algorithms to uncover similarities between items or users – and in the very interesting case where the sparsity is localized in a small fraction of the dataset – as in the *New Users Problem*, where new users are introduced to the system, they have not rated many items and thus, the CF algorithm can not make reliable personalized recommendations yet.

# *Ευχαριστίες*

Σε αυτό το σημείο, θα ήθελα να απευθύνω τις ιδιαίτερες ευχαριστίες μου σε όλους όσους συνέβαλλαν στην εκπόνηση της μεταπτυχιακής διπλωματικής μου εργασίας.

Αρχικά, θα ήθελα να ευχαριστήσω τον καθηγητή μου και επιβλέποντα, κ. Γιάννη Γαροφαλάκη για την πολύτιμη καθοδήγηση και βοήθεια. Επίσης, ευχαριστώ τους καθηγητές κ. Αθανάσιο Τσακαλίδη και κ. Ιωάννη Χατζηλυγερούδη που με τίμησαν και δέχτηκαν να συμμετάσχουν στην τριμελή επιτροπή αξιολόγησης της διπλωματικής μου.

Ιδιαίτερα, θα ήθελα να ευχαριστήσω το διδάκτορα Αθανάσιο Ν.Νικολακόπουλο για την πολύτιμη συμβολή και καθοδήγησή του σε όλη την πορεία των μεταπτυχιακών μου σπουδών. Τον ευχαριστώ ολόψυχα που με εμπιστεύτηκε και δέχτηκε να συνεργαστούμε για τη συγγραφή της κοινής μας δημοσίευσης. Η αστείρευτη βοήθεια, η επιστημονική του κατάρτιση και εμπειρία που μου προσέφερε απλόχερα συντέλεσαν στη δημιουργία της παρούσας διπλωματικής εργασίας.

Τέλος, οφείλω ένα μεγάλο ευχαριστώ στην οικογένειά μου και την αδελφική μου φίλη Δήμητρα, για την υποστήριξη, τη συμπαράσταση και την κατανόησή τους σε όλη τη διάρκεια των μεταπτυχιακών μου σπουδών. Μα πάνω απ' όλα, θα ήθελα να ευχαριστήσω μέσα απ' την καρδιά μου το σύζυγό μου Θάνο, που είναι πάντα δίπλα μου και με στηρίζει σε κάθε μου βήμα. Σ' ευχαριστώ ζωή μου ….

# Περιεχόμενα

# Κατάλογος σχημάτων

# Κατάλογος πινάκων

# Κεφάλαιο 1

# Εισαγωγή

## 1.1 Συστήματα Συστάσεων

Ο ολοένα αυξανόμενος όγκος online πληροφοριών που προκύπτει από τις ραγδαία εξελισσόμενες ηλεκτρονικές εφαρμογές και υπηρεσίες, έχει καταστήσει σαφή την ανάγκη ύπαρξης προσεγγίσεων οι οποίες είναι σε θέση να βοηθούν και να κατευθύνουν το χρήστη προς την άμεση και αποδοτική εξυπηρέτησή του, σχετικά με τις αποφάσεις που καλείται να πάρει.

Τα **Συστήματα Συστάσεων** (ΣΣ) είναι εργαλεία λογισμικού τα οποία παρέχουν αυτόματες και προσωποποιημένες συστάσεις στους χρήστες για *αντικείμενα* που πρόκειται να χρησιμοποιήσουν [45]. Οι συστάσεις έχουν να κάνουν με αποφάσεις των χρηστών σχετικά με το ποια αντικείμενα να αγοράσουν, ποια άρθρα να διαβάσουν, ποια ταινία να δουν, κλπ.

Ως *αντικείμενο* θεωρείται κάτι που προτείνει το σύστημα στους χρήστες. Συνήθως ένα ΣΣ επικεντρώνεται σε ένα είδος αντικειμένων (π.χ. ταινίες) και ολόκληρος ο σχεδιασμός του, από τη διεπαφή χρήστη μέχρι και την τεχνική συστάσεων που χρησιμοποιεί, έχει πραγματοποιηθεί με μόνο στόχο να παρέχει χρήσιμες και αποτελεσματικές συστάσεις για αυτό το συγκεκριμένο είδος αντικειμένων.

Προκειμένου ένα ΣΣ να πετύχει το σκοπό του και να αναγνωρίσει ενδιαφέροντα αντικείμενα για τον εκάστοτε χρήστη, θα πρέπει να *προβλέψει* ότι κάποιο αντικείμενο αξίζει να του το συστήσει. Για να το κάνει αυτό, το ΣΣ πρέπει να είναι ικανό να προβλέψει τη χρησιμότητα κάποιων αντικειμένων ή τουλάχιστον να συγκρίνει τη χρησιμότητα τους και μετά να αποφασίσει ποια από αυτά θα προτείνει με βάση αυτή τη σύγκριση.

Το πρόβλημα των συστάσεων μπορεί να οριστεί ως η εκτίμηση της απόκρισης του χρήστη για τα νέα αντικείμενα, με βάση παλαιότερες πληροφορίες που διαθέτει το σύστημα, και η πρόταση σε αυτόν το χρήστη καινούριων αντικειμένων για τα οποία η προβλεπόμενη απόκριση είναι υψηλή. Το είδος των αποκρίσεων χρηστών ποικίλει από εφαρμογή σε εφαρμογή. Για παράδειγμα, η απόκριση του χρήστη μπορεί να είναι αρέσει/δεν αρέσει, ενδιαφέρομαι/δεν ενδιαφέρομαι, ή βαθμολογία από 1 έως 5 για το πόσο του άρεσε κάτι με άριστα το 5. Από 'δω και πέρα, θα αναφερόμαστε στην απόκριση του χρήστη ως *βαθμολογία*.

Τα συστήματα συστάσεων βασίζονται σε διάφορα είδη εισόδου. Το πιο βολικό είναι η υψηλής ποιότητας άμεση ανατροφοδότηση (feedback), όπου οι χρήστες αναφέρουν απευθείας το ενδιαφέρον τους για αντικείμενα. Για παράδειγμα, το Netflix συλλέγει αστέρια ως βαθμολογία για ταινίες και οι χρήστες του TiVo εκφράζουν την προτίμησή τους για τηλεοπτικά σόου επιλέγοντας "thumbs up" και "thumbs down" για να δηλώσουν ότι τους αρέσει κάτι ή όχι αντίστοιχα. Ωστόσο, η άμεση ανατροφοδότηση δεν είναι πάντα διαθέσιμη, κι έτσι διάφορα συστήματα συμπεραίνουν τις προτιμήσεις των χρηστών από τις πιο εύκολα διαθέσιμες έμμεσες αποκρίσεις τους, οι οποίες αντικατοπτρίζουν γνώμες που έχουν προκύψει από τη μελέτη της συμπεριφοράς των χρηστών. Κάποια παραδείγματα έμμεσης ανατροφοδότησης είναι το ιστορικό αγορών, το ιστορικό περιήγησης, οι κινήσεις του ποντικιού, κ.λπ..

Δύο βασικά είδη ΣΣ είναι τα *Συνεργατικής Διήθησης (Collaborative Filtering, CF)* και τα *Διήθησης με Βάση το Περιεχόμενο (Content-based Filtering)*.

Οι CF τεχνικές θεωρούνται ως οι πιο διάσημες και ευρέως χρησιμοποιούμενες [45]. Σύμφωνα με την πιο απλή υλοποίηση αυτής της προσέγγισης, τα αντικείμενα που προτείνονται στο χρήστη είναι αυτά που άρεσαν κατά το παρελθόν σε άλλους χρήστες με παρόμοιες προτιμήσεις. Η ομοιότητα στις προτιμήσεις δύο χρηστών υπολογίζεται με βάση την ομοιότητα που υπάρχει στις βαθμολογίες που έχουν δώσει οι χρήστες. Αναφερόμαστε πιο αναλυτικά στο συγκεκριμένο είδος ΣΣ στο επόμενο κεφάλαιο της παρούσας εργασίας, καθώς ο αλγόριθμος που προτείνουμε ανήκει σε αυτήν την οικογένεια.

Στις τεχνικές διήθησης με βάση το περιεχόμενο, το ΣΣ μαθαίνει να προτείνει στους χρήστες αντικείμενα τα οποία είναι παρόμοια με αυτά που άρεσαν στο χρήστη κατά το παρελθόν. Τέτοιες μέθοδοι έχουν ως βασικό τους στόχο να αναγνωρίσουν τα κοινά χαρακτηριστικά εκείνων των αντικειμένων τα οποία έχουν βαθμολογηθεί θετικά από ένα χρήστη, και έπειτα να συστήσουν στον ίδιο χρήστη νέα αντικείμενα με τα ίδια χαρακτηριστικά. Για παράδειγμα, αν ένας χρήστης έχει βαθμολογήσει θετικά μία κωμωδία, τότε το σύστημα θα του προτείνει κι άλλες κωμωδίες. Με λίγα λόγια, η βασική διαδικασία που πραγματοποιείται από ένα σύστημα που βασίζεται στο περιεχόμενο είναι να ταιριάξει τις ιδιότητες ενός προφίλ χρήστη, οι οποίες φανερώνουν τις προτιμήσεις και τα ενδιαφέροντά του, με τις ιδιότητες ενός αντικειμένου, ώστε εντέλει να προτείνει στο χρήστη αντικείμενα που θα τον ενδιαφέρουν [34].

Συστήματα συστάσεων τα οποία βασίζονται αποκλειστικά στο περιεχόμενο αντιμετωπίζουν συνήθως προβλήματα όπως η *περιορισμένη ανάλυση* του περιεχομένου και η *υπερ-εξειδίκευση* [50]. Περιορισμένη ανάλυση του περιεχομένου προκύπτει από το γεγονός ότι το σύστημα μπορεί να έχει μόνο λίγες πληροφορίες για τους χρήστες του ή για το περιεχόμενο των αντικειμένων του. Αυτή η έλλειψη πληροφορίας μπορεί να οφείλεται σε διάφορες αιτίες. Για παράδειγμα, ζητήματα προσωπικών δεδομένων ενδεχομένως να προβληματίζουν ένα χρήστη με αποτέλεσμα να μην παρέχει προσωπικές του πληροφορίες, ή το ακριβές περιεχόμενο αντικειμένων μπορεί να είναι δύσκολο είτε κοστοβόρο να αποκτηθεί για κάποια είδη αντικειμένων όπως η μουσική ή οι φωτογραφίες. Επίσης, το περιεχόμενο ενός αντικειμένου είναι συχνά ανεπαρκές για να καθοριστεί η ποιότητά του. Παραδείγματος χάριν, ίσως είναι αδύνατο να γίνει διαχωρισμός ανάμεσα σε ένα καλώς γραμμένο και ένα κακώς γραμμένο άρθρο αν και τα δύο χρησιμοποιούν τους ίδιους όρους. Η υπερ-εξειδίκευση από την άλλη μεριά, είναι παρενέργεια του τρόπου με τον οποίο τα συστήματα με βάση το περιεχόμενο συστήνουν νέα αντικείμενα, όπου η προβλεπόμενη βαθμολογία ενός χρήστη για ένα αντικείμενο είναι υψηλή εάν αυτό το αντικείμενο είναι παρόμοιο με όσα άρεσαν στο χρήστη. Για παράδειγμα, σε μία εφαρμογή που συστήνει ταινίες, το σύστημα μπορεί να προτείνει σε ένα χρήστη μία ταινία της ίδιας κατηγορίας ή μία ταινία με τους ίδιους ηθοποιούς με ταινίες που έχει ήδη παρακολουθήσει αυτός ο χρήστης. Εξαιτίας αυτής της συμπεριφοράς, το σύστημα είναι πιθανό να μην μπορεί να συστήσει αντικείμενα τα οποία είναι διαφορετικά αλλά ταυτόχρονα ενδιαφέροντα στο χρήστη. Λύσεις που έχουν προταθεί για αυτό το πρόβλημα περιλαμβάνουν την προσθήκη τυχαιότητας [51] ή το φιλτράρισμα αντικειμένων τα οποία είναι ιδιαίτερα όμοια [6, 54].

Αξίζει να αναφέρουμε ένα ακόμα είδος ΣΣ, τα *Υβριδικά*. Αυτά τα συστήματα βασίζονται στο συνδυασμό τεχνικών. Ένα υβριδικό σύστημα που συνδυάζει δύο διαφορετικές τεχνικές, προσπαθεί να χρησιμοποιήσει τα πλεονεκτήματα της μιας για να επιλύσει τα μειονεκτήματα της άλλης. Για παράδειγμα, οι CF μέθοδοι δυσκολεύονται να διαχειριστούν τα προβλήματα που έχουν να κάνουν με τα νέα αντικείμενα, καθώς δεν μπορούν να πραγματοποιήσουν συστάσεις αντικειμένων τα οποία δεν έχουν καμία βαθμολογία από τους χρήστες. Κάτι τέτοιο ωστόσο, δεν αποτελεί περιορισμό για τα content-based συστήματα, αφού η πρόβλεψη για αυτά τα νέα αντικείμενα βασίζεται στην περιγραφή τους η οποία ενδεχομένως να είναι πιο εύκολα διαθέσιμη. Έχουν προταθεί διάφοροι τρόποι συνδυασμού δύο ή περισσότερων τεχνικών για τη δημιουργία ενός υβριδικού συστήματος.

## 1.2  Συνεισφορά της Εργασίας

Ο σκοπός της παρούσας διπλωματικής εργασίας είναι η μελέτη και ανάπτυξη μίας εναλλακτικής μεθόδου για το πρόβλημα των top-N συστάσεων σε συστήματα μεγάλου όγκου δεδομένων, η οποία να είναι υπολογιστικά αποδοτική και ταυτόχρονα να διατηρεί την ποιότητα συστάσεων, ακόμα και στις ιδιαίτερα δύσκολες συνθήκες όπου τα δεδομένα του συστήματος δεν επαρκούν για να πραγματοποιήσουν προσωποποιημένες συστάσεις (αραιότητα - sparsity).

Η κύρια συνεισφορά λοιπόν της συγκεκριμένης εργασίας είναι η πρόταση του **Lanczos Latent Factor Recommender** (LLFR), ενός νέου αλγόριθμου συνεργατικής διήθησης ο οποίος:

- Κατασκευάζει ένα μικρότερης διάστασης latent factor μοντέλο στο χώρο αντικειμένων, το οποίο ανακαλύπτει τις εγγενείς ομοιότητες ανάμεσα στα αντικείμενα, χρησιμοποιώντας μία προσέγγιση υπολογιστικά αποδοτική η οποία βασίζεται στη μέθοδο Lanczos. Κάτι τέτοιο, τον καθιστά μία οικονομική και επεκτάσιμη εναλλακτική, συγκριτικά με αντίστοιχους κοστοβόρους αλγόριθμους.

- Πετυχαίνει πολύ καλή ποιότητα συστάσεων σε ευρέως χρησιμοποιούμενες μετρικές, ξεπερνώντας διάφορες άλλες γνωστές και κοινά αποδεκτές τεχνικές.

- Παρουσιάζει μικρή ευαισθησία σε προβλήματα που προκαλούνται από την αραιότητα των δεδομένων, τα οποία αποτελούν μεγάλη πρόκληση για τις σύγχρονες και απαιτητικές εφαρμογές. Πιο συγκεκριμένα, ο LLFR αποδίδει καλύτερα τόσο στην περίπτωση όπου η αραιότητα είναι γενικευμένη, όπως στο *New Community Problem*, όσο και στην ιδιαίτερα ενδιαφέρουσα περίπτωση όπου η αραιότητα εντοπίζεται τοπικά σε ένα μικρό κομμάτι των δεδομένων, όπως στο *New Users Problem*.

Τα παραπάνω αποτελέσματα της παρούσας εργασίας έχουν δημοσιευτεί στο 4th International Conference on Web Intelligence Mining and Semantics, WIMS 2014, ACM, New York, USA, με τίτλο *On the Use of Lanczos Vectors for Efficient Latent Factor-Based Top-N Recommendation* [41]. Η παρουσίαση της εργασίας πραγματοποιήθηκε στο παραπάνω συνέδριο που έλαβε χώρα τον Ιούνιο του 2014 στη Θεσσαλονίκη.

## 1.3 Οργάνωση της Διπλωματικής

Η παρούσα διπλωματική εργασία οργανώνεται ως ακολούθως:

Στο Κεφάλαιο 2 γίνεται επισκόπηση της βιβλιογραφίας αναφορικά με τα Συστήματα Συστάσεων Συνεργατικής Διήθησης. Πιο συγκεκριμένα, αρχικά περιγράφουμε αναλυτικά τα χαρακτηριστικά αυτής της οικογένειας ΣΣ, ποια είναι τα βασικά είδη μεθόδων που ανήκουν σε αυτήν την κατηγορία και ποια προβλήματα αντιμετωπίζουν. Έπειτα, στην προσπάθεια να βρούμε έναν αποδοτικότερο τρόπο χειρισμού της αραιότητας των δεδομένων, μίας από τις μεγαλύτερες προκλήσεις που καλούνται να αντιμετωπίσουν σήμερα τέτοια συστήματα, παρουσιάζουμε οικογένειες μεθόδων οι οποίες θεωρείται ότι τα πηγαίνουν καλά σε τέτοιες περιπτώσεις όπου τα δεδομένα του συστήματος είναι αραιά: τα Latent Factor και Graph-Based μοντέλα.

Στο Κεφάλαιο 3 παρουσιάζουμε αναλυτικά τον LLFR Αλγόριθμο. Αρχικά, αναφερόμαστε στα latent factor μοντέλα και στον PureSVD, ο οποίος αποτελεί ένα από τα πιο πετυχημένα παραδείγματα για τις top-N συστάσεις. Στη συνέχεια, ξεκινώντας με μία σύντομη αναφορά στη μέθοδο Lanczos, περιγράφουμε λεπτομερώς το μοντέλο μας. Τέλος, παρουσιάζουμε τον ολοκληρωμένο αλγόριθμο καθώς και τα υπολογιστικά του χαρακτηριστικά.

Στο Κεφάλαιο 4 περιγράφουμε πλήρως την πειραματική διαδικασία που ακολουθήσαμε, τη μεθοδολογία αξιολόγησης της ποιότητας των συστάσεων που παράγονται και παρουσιάζουμε τα αποτελέσματα που προκύπτουν. Τέλος, εξετάζουμε το Πρόβλημα της Κρύας Εκκίνησης (Cold Start Problem) και ελέγχουμε την απόδοση του αλγορίθμου μας σε συνθήκες προσομοίωσης του συγκεκριμένου προβλήματος.

Στο Κεφάλαιο 5 συνοψίζουμε αναφέροντας τη συνεισφορά του LLFR και παρουσιάζουμε τα συμπεράσματά μας.

Τέλος, στο Παράρτημα Α παραθέτουμε τον κώδικα που χρησιμοποιήθηκε στη διπλωματική εργασία.

# Κεφάλαιο 2

# Συστήματα Συστάσεων Συνεργατικής Διήθησης

## 2.1 Συνεργατική Διήθηση - Collaborative Filtering

Αντίθετα με τις προσεγγίσεις με βάση το περιεχόμενο, οι οποίες για να παράξουν συστάσεις χρησιμοποιούν το περιεχόμενο των αντικειμένων εκείνων που έχουν βαθμολογηθεί στο παρελθόν από ένα συγκεκριμένο χρήστη, οι προσεγγίσεις συνεργατικής διήθησης βασίζονται μεν στις βαθμολογίες του συγκεκριμένου χρήστη αλλά βασίζονται και σε αυτές από άλλους χρήστες. Η βασική ιδέα είναι ότι η βαθμολογία ενός χρήστη $u$ για ένα νέο αντικείμενο $i$ είναι πιθανό να είναι παρόμοια με αυτή ενός άλλου χρήστη $v$, αν οι $u$ και $v$ έχουν βαθμολογήσει άλλα αντικείμενα με παρόμοιο τρόπο. Παρόμοια, ο $u$ είναι πιθανό να βαθμολογήσει δύο αντικείμενα $i$ και $j$ με παρόμοιο τρόπο, αν άλλοι χρήστες έχουν δώσει παρόμοιες βαθμολογίες σε αυτά τα δύο αντικείμενα.

Οι τεχνικές συνεργατικής διήθησης καταφέρνουν να ξεπερνούν κάποιους από τους περιορισμούς που αντιμετωπίζουν εκείνες που βασίζονται στο περιεχόμενο. Για παράδειγμα, αντικείμενα για τα οποία το σύστημα δεν έχει διαθέσιμο περιεχόμενο ή είναι δύσκολο να αποκτήσει, μπορούν και πάλι να προταθούν σε χρήστες μέσω απαντήσεων από άλλους χρήστες. Επιπλέον, οι συνεργατικές συστάσεις βγάζουν συμπεράσματα για την ποιότητα των αντικειμένων με βάση την αξιολόγηση που προκύπτει από τους χρήστες, αντί να βασίζονται στο περιεχόμενο το οποίο μπορεί να είναι κακός δείκτης ποιότητας. Τέλος, οι συνεργατικές μέθοδοι, σε αντίθεση με τα συστήματα που βασίζονται στο περιεχόμενο, μπορούν να προτείνουν αντικείμενα με πολύ διαφορετικό περιεχόμενο, αρκεί άλλοι χρήστες να έχουν δείξει ενδιαφέρον για αυτά τα διαφορετικά αντικείμενα.

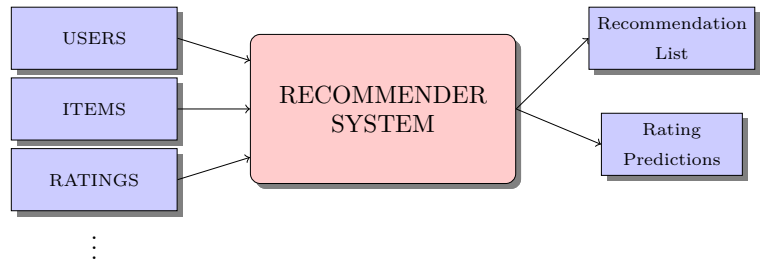

Σχήμα 2.1: Παράδειγμα ενός Συστήματος Συστάσεων [40].

Δοθέντος ενός συνόλου χρηστών, ενός συνόλου αντικειμένων και βαθμολογιών - οι οποίες έχουν τεθεί ρητά ή υπονοούνται – σχετικά με το πόσο αρέσουν ή όχι σε ένα χρήστη τα αντικείμενα που έχει ήδη δει (βλ. Σχήμα 2.1), οι παραδοσιακές CF τεχνικές προσπαθούν να δημιουργήσουν "γειτονιές", με βάση τις ομοιότητες ανάμεσα στους χρήστες (*user-oriented* CF) ή τα αντικείμενα (*item-oriented* CF) όπως αυτές προκύπτουν από τα δεδομένα. Ο στόχος τους είναι είτε να προβλέψουν τα σκορ προτίμησης για τα άγνωστα ζευγάρια χρήστη-αντικειμένου (*prediction-based* recommendation), είτε να δημιουργήσουν μία λίστα με αντικείμενα τα οποία είναι πιθανό να ενδιαφέρουν το χρήστη (*ranking-based* or *top-N* recommendation). Συνοψίζοντας, η διαδικασία που ακολουθούν οι μέθοδοι συνεργατικής διήθησης για την παραγωγή συστάσεων παρουσιάζεται στο Σχήμα 2.2.

Προκειμένου να διατυπώσουν συστάσεις, τα συστήματα CF χρειάζεται να συσχετίσουν δύο ριζικά διαφορετικές οντότητες: αντικείμενα και χρήστες. Υπάρχουν δυο βασικές προσεγγίσεις για την πραγματοποίηση αυτής της σύγκρισης, οι οποίες αποτελούν και τις δύο βασικές τεχνικές του CF: η *neighborhood* προσέγγιση (ή memory-based) και οι *model-based* προσεγγίσεις. Οι neighborhood μέθοδοι επικεντρώνονται στις σχέσεις ανάμεσα στα αντικείμενα ή, εναλλακτικά, ανάμεσα στους χρήστες. Οι model-based προσεγγίσεις (ή latent factor μοντέλα), όπως η παραγοντοποίηση μητρώων, περιλαμβάνουν μία εναλλακτική προσέγγιση προβάλλοντας και τα αντικείμενα και τους χρήστες στον ίδιο λανθάνοντα χώρο. Ο χώρος αυτός προσπαθεί να εξηγήσει τις βαθμολογίες χαρακτηρίζοντας και τα αντικείμενα και τους χρήστες με βάση "*παράγοντες*" που προκύπτουν αυτόματα από την αλληλεπίδραση του χρήστη.

Στις **neighborhood-based** CF μεθόδους, οι βαθμολογίες των χρηστών για τα αντικείμενα που διαθέτει το σύστημα χρησιμοποιούνται απευθείας για την πρόβλεψη βαθμολογιών για νέα αντικείμενα. Αυτό μπορεί να γίνει με δύο τρόπους οι οποίοι είναι γνωστοί ως user-based ή item-based συστάσεις. Η αρχική μορφή των μεθόδων που ανήκουν σε αυτήν την οικογένεια, την οποία είχαν υιοθετήσει όλα τα αρχικά CF συστήματα, βασίζεται στη user-based προσέγγιση. Τέτοιες μέθοδοι υπολογίζουν τις άγνωστες βαθμολογίες με βάση τις καταχωρημένες βαθμολογίες χρηστών με παρόμοια γούστα. Αργότερα, έγινε διάσημη μία ανάλογη item-based προσέγγιση. Σε αυτές τις μεθόδους, μία βαθμολογία εκτιμάται χρησιμοποιώντας γνωστές βαθμολογίες από τον ίδιο χρήστη αλλά σε παρόμοια αντικείμενα.

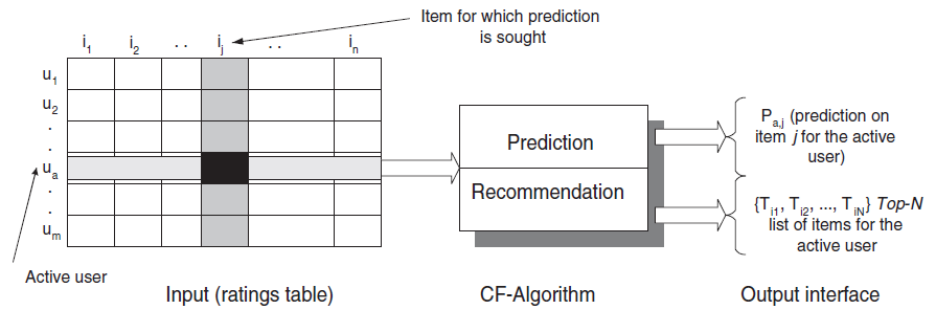

ΣΧΗΜΑ 2.2: Διαδικασία παραγωγής συστάσεων μέσω συνεργατικής διήθησης [47].

Για παράδειγμα, μπορούμε να δούμε τον Πίνακα 2.1, στον οποίο παρουσιάζονται οι βαθμολογίες πέντε χρηστών σε πέντε ταινίες. Ο Αλέξης πρέπει να αποφασίσει αν θα νοικιάσει την ταινία "Εκδικητές" την οποία δεν έχει δει.

ΠΙΝΑΚΑΣ 2.1: Παράδειγμα ενός μητρώου χρήστη-αντικειμένου (μητρώο βαθμολογιών). Το σύστημα πρέπει να κάνει συστάσεις για τον Αλέξη. Κάποιοι χρήστες δεν έχουν δώσει βαθμολογία σε κάποιες ταινίες, διότι δεν τις έχουν δει ακόμα.

|  | Ροζ Πάνθηρας | Εκδικητές | Μπάτμαν | Τιτανικός | Μονομάχος |
|---|---|---|---|---|---|
| Μαρία | 2 | 3 |  | 5 | 2 |
| Γιάννης | 1 | 5 | 3 | 2 |  |
| Αλέξης | 5 | ? | 3 | 3 | 1 |
| Άννα | 3 | 2 | 4 |  | 2 |
| Γιώργος |  | 2 | 4 | 3 |  |

Σε αντίθεση με τα neighborhood-based συστήματα, τα οποία χρησιμοποιούν απευθείας τις βαθμολογίες στην πρόβλεψη, οι **model-based** προσεγγίσεις τις χρησιμοποιούν για να μάθουν ένα μοντέλο προβλέψεων. Η γενική ιδέα είναι η μοντελοποίηση των αλληλεπιδράσεων χρήστη-αντικειμένου με παράγοντες οι οποίοι εκπροσωπούν λανθάνοντα χαρακτηριστικά των χρηστών και των αντικειμένων στο σύστημα, όπως η κλάση προτιμήσεων των χρηστών και η κλάση κατηγοριών των αντικειμένων. Αυτό το μοντέλο στη συνέχεια εκπαιδεύεται χρησιμοποιώντας τα διαθέσιμα δεδομένα, και στη συνέχεια χρησιμοποιείται για την πρόβλεψη βαθμολογιών χρηστών για νέα αντικείμενα. Για το πρόβλημα της σύστασης αντικειμένων, υπάρχουν πολλές προσεγγίσεις αυτής της οικογένειας και κάποιες από αυτές είναι οι Latent Semantic Analysis, Maximum Entropy, και Singular Value Decomposition.

Οι παραδοσιακές τεχνικές συνεργατικής διήθησης μπορούν να ομαδοποιηθούν στις ακόλουθες δύο κατηγορίες ανάλογα με τα αποτελέσματα που παράγουν:

**Prediction-based Recommendation.** Οι μέθοδοι αυτές προσπαθούν να μαντέψουν τη βαθμολογία προτίμησης του χρήστη για κάποιο συγκεκριμένο αντικείμενο, και με βάση αυτή να προχωρήσουν στη σύσταση αντικειμένων. Η επιτυχία των αποτελεσμάτων που παράγουν υπολογίζεται από μετρικές οι οποίες ελέγχουν την απόσταση ανάμεσα στις πραγματικές βαθμολογίες και στις προβλεπόμενες.

**Ranking-based Recommendation.** Ο στόχος των τεχνικών αυτών είναι η παραγωγή μίας λίστας με $N$ αντικείμενα τα οποία αναμένεται να ενδιαφέρουν το χρήστη (ranking-based ή top-N συστάσεις). Η λίστα αυτή κατατάσσει τα αντικείμενα με φθίνουσα σειρά προτίμησης του χρήστη. Με άλλα λόγια, σε αυτές τις περιπτώσεις η μέθοδος δε χρειάζεται να μαντέψει την **ακριβή βαθμολογία** που θα έδινε ο χρήστης αν χαρακτήριζε κάποιο αντικείμενο, αλλά αρκεί να το συμπεριλάβει στη λίστα και να το τοποθετήσει στη σωστή θέση μέσα σε αυτή, συγκριτικά με τα άλλα αντικείμενα που εμπεριέχονται. Αυτές οι μέθοδοι βρίσκουν ολοένα και μεγαλύτερη εφαρμογή στα σύγχρονα εμπορικά συστήματα, καθώς αυτά αρκούνται στο να εμφανίσουν στο χρήστη μερικά συγκεκριμένα αντικείμενα τα οποία προβλέπεται να τον ενδιαφέρουν περισσότερο, ενώ δε χρειάζεται να εμφανίσουν τις βαθμολογίες αξιολόγησης.

### 2.1.1 Matrix Factorization Models

Τα Latent factor models προσεγγίζουν τη συνεργατική διήθηση με στόχο να αποκαλύψουν τα λανθάνοντα χαρακτηριστικά που δικαιολογούν τις παρατηρούμενες βαθμολογίες. Παραδείγματα αποτελούν τα νευρωνικά δίκτυα και μοντέλα τα οποία προκύπτουν από την παραγοντοποίηση του μητρώου βαθμολογιών χρήστη-αντικειμένου (είναι επίσης γνωστά και ως μοντέλα που βασίζονται στον SVD). Τελευταία, τα μοντέλα παραγοντοποίησης μητρώου έχουν γίνει γνωστά, χάρη στα ελκυστικά χαρακτηριστικά ακρίβειας και επεκτασιμότητας.

Στην ανάκτηση πληροφορίας, ο SVD είναι καθιερωμένος για την αναγνώριση λανθανόντων σημασιολογικών παραγόντων [15]. Ωστόσο, στο CF η εφαρμογή του SVD στις άμεσες βαθμολογίες δημιουργεί δυσκολίες εξαιτίας του μεγάλου ποσοστού τιμών που λείπουν. Ο συμβατικός SVD δεν ορίζεται όταν λείπουν τιμές από το μητρώο. Επιπλέον, ο μη προσεκτικός χειρισμός των σχετικά ελάχιστων γνωστών εισόδων τον καθιστά ιδιαίτερα επιρρεπή στο *overfitting*. Στην προσπάθειά του να "ταιριάξει" όσο καλύτερα μπορεί τα πραγματικά δεδομένα, καταλήγει να το "παρακάνει" (over-fit) και να κινδυνεύει να συμπεριλάβει και το θόρυβο (στην περίπτωση που εξετάζουμε πρόκειται για σκουπίδια/ψεύτικη πληροφορία στα πραγματικά δεδομένα) μέσα στα πραγματικά δεδομένα. Το αποτέλεσμα φυσικά είναι να παράγονται λανθασμένες προβλέψεις, εφόσον ο θόρυβος έχει παραμορφώσει τα δεδομένα.

Προηγούμενες εργασίες βασίστηκαν στο *imputation* [47], κατά το οποίο συμπληρώνονται οι βαθμολογίες που λείπουν με κάποια τιμή και το μητρώο βαθμολογιών πυκνώνει. Ωστόσο, το imputation μπορεί να γίνει πολύ ακριβό καθώς αυξάνει πολύ την ποσότητα των δεδομένων. Επιπρόσθετα, τα δεδομένα μπορεί να παραμορφωθούν σημαντικά λόγω της προσέγγισης με μη ακριβείς τιμές. Έτσι, πιο πρόσφατες εργασίες [4, 10, 29, 52] πρότειναν τη μοντελοποίηση απευθείας μόνο των παρατηρούμενων βαθμολογιών, ενώ ταυτόχρονα αποφεύγουν το overfitting μέσα από ένα επαρκώς regularized μοντέλο.

Στη βασική της μορφή, η παραγοντοποίηση μητρώου χαρακτηρίζει και τα αντικείμενα και τους χρήστες με διανύσματα παραγόντων που προκύπτουν από μοτίβα μέσα από τις βαθμολογίες των αντικειμένων. Υψηλή αντιστοιχία ανάμεσα στους παράγοντες αντικειμένου και χρήστη οδηγεί στη σύσταση ενός αντικειμένου σε ένα χρήστη. Αυτές οι μέθοδοι καταφέρνουν ακρίβεια πρόβλεψης μεγαλύτερη από άλλες δημοσιευμένες τεχνικές συνεργατικής διήθησης. Ταυτόχρονα, προσφέρουν ένα αποδοτικό από άποψη μνήμης μοντέλο, το οποίο μπορεί σχετικά εύκολα να εκπαιδευτεί. Αυτό που κάνει αυτές τις τεχνικές ακόμα πιο βολικές είναι η ικανότητά τους να διαχειρίζονται διάφορα κρίσιμα θέματα που έχουν να κάνουν με τα δεδομένα. Αρχικά, έχουν τη δυνατότητα να μπορούν να ανταπεξέρχονται σε πολλαπλά είδη ανατροφοδότησης του χρήστη (συγκεκριμένη βαθμολογία, χαρακτηρισμός "μου αρέσει/δε μου αρέσει", κλπ.). Έπειτα, μπορούν να προβλέψουν καλύτερα τις βαθμολογίες χρήστη παρατηρώντας επίσης άλλες σχετικές ενέργειες του ίδιου χρήστη, όπως το ιστορικό αγορών και πλοήγησης [30].

### 2.1.2 Μοντέλα Γειτνίασης - Neighborhood Models

Η πιο συχνή προσέγγιση στη συνεργατική διήθηση βασίζεται στα μοντέλα γειτνίασης. Οι αλγόριθμοι αυτής της οικογένειας για να παράξουν μία πρόβλεψη χρησιμοποιούν ολόκληρη ή ένα μέρος της πληροφορίας από το μητρώο βαθμολογιών. Ο στόχος τους είναι να εντοπίσουν τους "γείτονες" ενός νέου χρήστη. Πιο συγκεκριμένα, για να πραγματοποιήσουν μία σύσταση υπολογίζουν την ομοιότητα ή το βάρος μεταξύ δύο χρηστών ή δύο αντικειμένων, και υπολογίζουν το σταθμισμένο μέσο όρο όλων των βαθμολογιών που έχει δώσει ένας συγκεκριμένος χρήστης ή που έχει πάρει ένα συγκεκριμένο αντικείμενο [47]. Η παραπάνω διαδικασία υφίσταται για την παραγωγή προτεινόμενων βαθμολογιών (prediction-based recommendation). Για την περίπτωση όπου παράγεται μία λίστα top-N συστάσεων, τότε οι παραπάνω αλγόριθμοι μέσα από τον υπολογισμό των ομοιοτήτων εντοπίζουν τους $k$ πιο όμοιους χρήστες ή αντικείμενα ($k$ nearest neighbors), και τους συγκεντρώνουν ώστε να πάρουν τα $N$ πιο συχνά αντικείμενα τα οποία αποτελούν και τη σύσταση.

**User-based συστάσεις** Οι user-based μέθοδοι συστάσεων προβλέπουν μία βαθμολογία $r_{ui}$ ενός χρήστη $u$ για ένα αντικείμενο $i$ χρησιμοποιώντας τις βαθμολογίες που έχουν δοθεί στο $i$ από χρήστες περισσότερο όμοιους (παρουσιάζουν ομοιότητες στις βαθμολογίες) στον $u$, οι οποίοι ονομάζονται *κοντινότεροι γείτονες*. Οι γείτονες του χρήστη $u$ είναι ουσιαστικά οι χρήστες $v$ των οποίων οι βαθμολογίες στα αντικείμενα που έχουν βαθμολογηθεί και από τον $u$ και από τους $v$ ταιριάζουν πιο πολύ σε αυτές του $u$. Αξίζει να αναφέρουμε ότι διάφοροι παράμετροι πρέπει να λαμβάνονται υπόψη σε προσεγγίσεις αυτής της οικογένειας, όπως το ότι οι χρήστες δε βαθμολογούν με τον ίδιο τρόπο αντικείμενα που τους αρέσουν (άλλοι είναι πιο αυστηροί στη βαθμολογία τους), ή το ότι ανάμεσα στους γείτονες ενός χρήστη κάποιοι μπορεί να είναι πιο κοντά του σε σχέση με κάποιους άλλους (δίνονται βάρη στη συνεισφορά του κάθε γείτονα, ανάλογα με το βαθμό ομοιότητάς του με το χρήστη). Παραδείγματα user-based συστημάτων είναι τα GroupLens [27] και Ringo[50].

**Item-based συστάσεις** Από την άλλη μεριά, ενώ οι user-based μέθοδοι βασίζονται στην άποψη χρηστών με παρόμοια γούστα, οι item-based προσεγγίσεις [16, 33] εξετάζουν τις βαθμολογίες που έχουν δοθεί σε όμοια αντικείμενα. Πιο συγκεκριμένα, η βαθμολογία $r_{ui}$ ενός χρήστη $u$ για ένα αντικείμενο $i$ προβλέπεται με βάση τις βαθμολογίες που έχει δώσει ο χρήστης σε άλλα παρόμοια αντικείμενα. Σε τέτοιες προσεγγίσεις, δύο αντικείμενα είναι παρόμοια αν τα έχουν βαθμολογήσει με παρόμοιο τρόπο αρκετοί χρήστες του συστήματος. Και σε αυτήν την οικογένεια προσεγγίσεων απαιτείται κανονικοποίηση των βαθμολογιών, προκειμένου για παράδειγμα και εδώ να ληφθούν υπόψη οι διαφορετικές βαθμολογικές κλίμακες των χρηστών. Οι item-based προσεγγίσεις κέρδισαν αμέσως έδαφος σε πολλές περιπτώσεις χάρη στην καλύτερη επεκτασιμότητα και τη βελτιωμένη ακρίβειά τους συγκριτικά με τις user-based μεθόδους. Επιπλέον, οι item-based μέθοδοι δίνουν μια πιο κατανοητή εξήγηση για τις προβλέψεις τους. Αυτό ισχύει διότι οι χρήστες είναι εξοικειωμένοι με αντικείμενα τα οποία είχαν προτιμήσει στο παρελθόν, αλλά δε γνωρίζουν εκείνα τα οποία υποτίθεται ότι συστήνονται λόγω άλλων χρηστών με παρόμοια γούστα.

#### 2.1.2.1 Πλεονεκτήματα των Μοντέλων Γειτνίασης

Ενώ πρόσφατες έρευνες δείχνουν ότι μοντέρνες model-based προσεγγίσεις είναι ανώτερες από τις neighborhood για την πρόβλεψη βαθμολογιών [29, 52], γίνεται αντιληπτό ότι η καλή ακρίβεια πρόβλεψης από μόνη της δεν εγγυάται στους χρήστες μία αποτελεσματική και ικανοποιητική εμπειρία [22]. Προς αυτήν την κατεύθυνση, ένας άλλος παράγοντας ο οποίος φαίνεται να παίζει σημαντικό ρόλο στο κατά πόσο οι χρήστες εκτιμούν το σύστημα συστάσεων είναι το *serendipity* (απρόσμενα ευχάριστη έκπληξη) [22]. Πρόκειται για κάτι καινούριο το οποίο βρίσκει ενδιαφέρον ο χρήστης και που διαφορετικά μπορεί να μην ανακάλυπτε. Για παράδειγμα, η σύσταση σε ένα χρήστη μιας ταινίας που έχει σκηνοθετήσει ο αγαπημένος του σκηνοθέτης, αποτελεί μια νέα σύσταση αν ο χρήστης δε γνώριζε την ύπαρξη αυτής της ταινίας, αλλά πιθανότατα θα την ανακάλυπτε μόνος του. Κάτι τέτοιο δεν αποτελεί serendipity.

Οι model-based προσεγγίσεις το παρακάνουν στο να χαρακτηρίζουν τις προτιμήσεις ενός χρήστη με latent factors. Για παράδειγμα, σε ένα σύστημα που συστήνει ταινίες, τέτοιες μέθοδοι μπορεί να προσδιορίσουν ότι σε ένα χρήστη αρέσουν ταινίες οι οποίες είναι και αστείες και ρομαντικές, χωρίς να χρειάζεται να ορίσουν τις έννοιες "αστείο" και "ρομαντικό". Ένα τέτοιο σύστημα είναι σε θέση να συστήσει σε ένα χρήστη μία ρομαντική κωμωδία που δε γνώριζε ο χρήστης. Ωστόσο, ενδεχομένως να είναι δύσκολο για αυτό το σύστημα να συστήσει μία ταινία η οποία δεν ταιριάζει ακριβώς με αυτήν την υψηλού επιπέδου κατηγορία. Οι neighborhood προσεγγίσεις από την άλλη μεριά, συλλαμβάνουν τοπικές συσχετίσεις στα δεδομένα. Συνεπώς, είναι πιθανό για ένα σύστημα συστάσεων ταινιών που χρησιμοποιεί μια τέτοια τεχνική να συστήσει στο χρήστη μία ταινία διαφορετική από τα συνηθισμένα γούστα του ή μία ταινία η οποία δεν είναι ιδιαίτερα γνωστή, αν κάποιος από τους στενούς γείτονές του της έχει δώσει υψηλή βαθμολογία. Μία τέτοια σύσταση μπορεί να μην είναι εγγυημένα επιτυχής, όπως θα ήταν μία ρομαντική κομεντί, αλλά ίσως βοηθήσει το χρήστη να ανακαλύψει μία τελείως νέα κατηγορία ή έναν νέο αγαπημένο ηθοποιό/σκηνοθέτη.

Γενικά, οι model-based τεχνικές καταφέρνουν να εκφράσουν ιδιαίτερα καλά τις διάφορες εκδοχές των δεδομένων. Έτσι, τείνουν να παρέχουν πιο ακριβή αποτελέσματα από τα μοντέλα γειτνίασης. Ωστόσο, τα περισσότερα εμπορικά συστήματα (για παράδειγμα, τα Amazon [33] και TiVo [2]) βασίζονται στα μοντέλα γειτνίασης. Αυτή τους η κυριαρχία οφείλεται εν μέρει στην απλότητά τους. Ωστόσο, υπάρχουν πιο σημαντικοί λόγοι, όπως θα δούμε αμέσως παρακάτω, για να παραμένουν πιστά σε αυτά τα μοντέλα αρκετά πραγματικά συστήματα.

Τα κύρια πλεονεκτήματα των neighborhood-based μεθόδων είναι [17]:

**Απλότητα.** Τέτοιες μέθοδοι είναι διαισθητικές και σχετικά απλές στην εφαρμογή τους. Στην πιο απλή τους μορφή, μόνο μία παράμετρος (ο αριθμός των γειτόνων που θα χρησιμοποιηθούν στην πρόβλεψη) χρειάζεται να ρυθμιστεί.

**Δυνατότητα αιτιολόγησης.** Οι neighborhood-based μέθοδοι προσφέρουν επίσης μία σύντομη και διαισθητική αιτιολόγηση για τις συστάσεις που πραγματοποιούνται. Για παράδειγμα, στις item-based συστάσεις, η λίστα με τα γειτονικά αντικείμενα, καθώς και οι βαθμολογίες που έχουν δοθεί από το χρήστη για αυτά τα αντικείμενα, μπορούν να παρουσιαστούν στο χρήστη ως αιτιολόγηση για τη σύσταση που του έχει γίνει. Κάτι τέτοιο μπορεί να βοηθήσει το χρήστη να κατανοήσει καλύτερα τις συστάσεις και τη συνάφειά τους, και θα μπορούσε να λειτουργήσει ως βάση για ένα αλληλεπιδραστικό σύστημα στο οποίο οι χρήστες μπορούν να επιλέξουν τους γείτονες για τους οποίους θα δοθεί μεγαλύτερη σημασία στη σύσταση [4].

**Αποτελεσματικότητα.** Ένα από τα πιο δυνατά σημεία των neighborhood-based συστημάτων είναι η αποτελεσματικότητά τους. Σε αντίθεση με τα model-based συστήματα, δεν απαιτούν εκπαίδευση, μία απαίτηση η οποία πρέπει να πραγματοποιείται σε τακτά χρονικά διαστήματα σε μεγάλες εμπορικές εφαρμογές. Ενώ η διαδικασία των συστάσεων είναι συνήθως πιο ακριβή όταν πρόκειται για model-based μεθόδους, οι nearest-neighbors μπορούν να υπολογισθούν από πριν σε ένα off-line βήμα, προσφέροντας σχεδόν στιγμιαίες συστάσεις. Επιπλέον, η αποθήκευση αυτών των nearest-neighbors έχει μικρές απαιτήσεις σε μνήμη, καθιστώντας τέτοιες προσεγγίσεις επεκτάσιμες σε εφαρμογές με εκατομμύρια χρηστών και αντικειμένων.

**Σταθερότητα.** Μια ακόμα χρήσιμη ιδιότητα των συστημάτων συστάσεων που βασίζονται σε αυτήν την προσέγγιση είναι ότι επηρεάζονται ελάχιστα από τη διαρκή προσθήκη χρηστών, αντικειμένων και βαθμολογιών, κάτι το οποίο παρατηρείται κατά κανόνα σε μεγάλες εμπορικές εφαρμογές. Για παράδειγμα, ένα item-based σύστημα είναι σε θέση να κάνει συστάσεις σε νέους χρήστες αμέσως μόλις υπολογιστούν οι ομοιότητες ανάμεσα στα αντικείμενα, χωρίς να χρειάζεται να ξανά-εκπαιδεύσει το σύστημα. Επιπλέον, μόλις εισαχθούν κάποιες βαθμολογίες για ένα νέο αντικείμενο, το μόνο που απαιτείται να υπολογιστεί είναι οι ομοιότητες ανάμεσα σε αυτό το αντικείμενο και σε αυτά που υπάρχουν ήδη στο σύστημα.

### 2.1.2.2 Μειονεκτήματα των Μοντέλων Γειτνίασης

Οι προσεγγίσεις γειτνίασης που βασίζονται στις συσχετίσεις ανάμεσα στις βαθμολογίες παρουσιάζουν δύο σημαντικά μειονεκτήματα:

**Περιορισμένη κάλυψη (limited coverage).** Η κάλυψη αφορά στο εύρος αντικειμένων που μπορεί να συστήσει ένα ΣΣ. Επειδή οι συσχετίσεις μεταξύ βαθμολογιών μετρούν την ομοιότητα ανάμεσα σε δύο χρήστες συγκρίνοντας τις βαθμολογίες τους στα ίδια αντικείμενα, οι χρήστες μπορούν να είναι γείτονες μόνο εάν έχουν βαθμολογήσει κοινά αντικείμενα. Αυτή η υπόθεση είναι πολύ περιοριστική, καθώς χρήστες οι οποίοι έχουν βαθμολογήσει λίγα ή και καθόλου κοινά αντικείμενα μπορεί και πάλι να έχουν παρόμοιες προτιμήσεις. Επιπλέον, αφού μπορούν να προταθούν μόνο αντικείμενα τα οποία έχουν βαθμολογηθεί από γείτονες, η κάλυψη τέτοιων μεθόδων μπορεί επίσης να είναι περιορισμένη.

**Ευαισθησία στα αραιά δεδομένα.** Οι μέθοδοι συστάσεων που βασίζονται στη γειτονικότητα υποφέρουν επίσης από την έλλειψη διαθέσιμων βαθμολογιών. Η αραιότητα είναι ένα κοινό πρόβλημα για τα περισσότερα συστήματα συστάσεων [8, 38, 40, 43, 44] λόγω του ότι οι χρήστες συνήθως βαθμολογούν μόνο ένα μικρό μέρος των διαθέσιμων αντικειμένων [5, 22, 46]. Στο μεταξύ, αυτό ενισχύεται και από το γεγονός ότι οι νέοι χρήστες ή τα νέα αντικείμενα που έχουν προστεθεί σε ένα σύστημα είναι πιθανό να μην έχουν καθόλου βαθμολογίες. Αυτό το πρόβλημα είναι γνωστό ως *Το Πρόβλημα της Κρύας Εκκίνησης* [48]. Όταν τα δεδομένα είναι αραιά, δύο χρήστες ή αντικείμενα είναι απίθανο να έχουν κοινές βαθμολογίες, και κατά συνέπεια, οι προσεγγίσεις που βασίζονται στη γειτονικότητα θα προβλέψουν βαθμολογίες χρησιμοποιώντας έναν πολύ περιορισμένο αριθμό από γείτονες. Επιπλέον, οι βαθμοί ομοιότητας μπορεί να υπολογίζονται χρησιμοποιώντας μόνο ένα μικρό αριθμό βαθμολογιών, οδηγώντας έτσι σε "προκατειλημμένες" συστάσεις.

Μία κοινή λύση για όλα αυτά τα προβλήματα είναι η συμπλήρωση των βαθμολογιών που λείπουν με κάποιες προκαθορισμένες τιμές [16], όπως η μέση τιμή του εύρους βαθμολογιών, και η μέση βαθμολογία. Μια πιο αξιόπιστη προσέγγιση είναι να χρησιμοποιηθεί πληροφορία από το περιεχόμενο για τη συμπλήρωση των βαθμολογιών που λείπουν [22, 27]. Για παράδειγμα, τα δεδομένα που λείπουν μπορούν να παραχθούν από αυτόνομους πράκτορες, γνωστούς ως *filterbots* [22, 27], οι οποίοι ενεργούν ως συνηθισμένοι χρήστες του συστήματος και βαθμολογούν αντικείμενα με βάση κάποια συγκεκριμένα χαρακτηριστικά του περιεχομένου τους. Από την άλλη μεριά, τα δεδομένα που λείπουν μπορούν να προβλεφθούν και από μία προσέγγιση με βάση το περιεχόμενο [36]. Ωστόσο, και αυτές οι λύσεις έχουν τα μειονεκτήματά τους. Για παράδειγμα, το να δώσει κανείς μία προκαθορισμένη τιμή στις βαθμολογίες που λείπουν μπορεί να προκαλέσει προκατάληψη στις συστάσεις. Επίσης, μπορεί να μην είναι διαθέσιμο περιεχόμενο σχετικά με τα αντικείμενα για τον υπολογισμό βαθμολογιών ή ομοιοτήτων.

Στις επόμενες ενότητες παρουσιάζονται δύο από τις βασικότερες προσεγγίσεις για την αντιμετώπιση των προβλημάτων της περιορισμένης κάλυψης και της αραιότητας: οι dimensionality reduction και οι graph-based μέθοδοι.

## 2.2 Latent Factor Models

Οι dimensionality reduction μέθοδοι [4, 5, 29, 42, 46, 53] αντιμετωπίζουν τα προβλήματα μειωμένης κάλυψης και αραιότητας προβάλλοντας χρήστες και αντικείμενα σε ένα μειωμένης διάστασης λανθάνοντα χώρο ο οποίος συλλαμβάνει τα πιο βασικά τους χαρακτηριστικά. Λόγω του ότι χρήστες και αντικείμενα συγκρίνονται σε αυτόν τον πυκνό υποχώρο με υψηλού επιπέδου χαρακτηριστικά αντί για τον χώρο βαθμολογιών, μπορούν να ανακαλυφθούν σχέσεις με μεγαλύτερο νόημα. Πιο συγκεκριμένα, μπορεί να βρεθεί μία σχέση ανάμεσα σε δύο χρήστες, ακόμα και αν αυτοί οι χρήστες έχουν βαθμολογήσει διαφορετικά αντικείμενα. Ως αποτέλεσμα, τέτοιες μέθοδοι είναι γενικά λιγότερο ευαίσθητες στα αραιά δεδομένα [4, 5, 46].

Κατά κύριο λόγο, υπάρχουν δύο τρόπους τους οποίους μπορούν να χρησιμοποιήσουν οι μέθοδοι αυτής της κατηγορίας: η διάσπαση του μητρώου βαθμολογιών, και η διάσπαση ενός αραιού μητρώου ομοιοτήτων.

### 2.2.1 Διάσπαση του μητρώου βαθμολογιών

Μια διάσημη dimensionality reduction προσέγγιση για τη σύσταση αντικειμένων είναι η *Latent Semantic Indexing (LSI)* [15]. Σε αυτήν την προσέγγιση, το $|\mathbf{U}| \times |\mathbf{I}|$ χρήστη-αντικείμενο μητρώο βαθμολογιών $\mathbf{R}$ τάξης $n$ προσεγγίζεται από ένα μητρώο $\hat{\mathbf{R}} = \mathbf{PQ}^\top$ τάξης $k < n$, όπου $\mathbf{P}$ είναι ένα $|\mathbf{U}| \times k$ μητρώο από παράγοντες σχετικούς με χρήστες και $\mathbf{Q}$ ένα $|\mathbf{I}| \times k$ μητρώο από παράγοντες αντικειμένων. Διαισθητικά, η $u$-οστή γραμμή του $\mathbf{P}$, $\mathbf{p_u} \in \mathbb{R}^k$, αναπαριστά τις συντεταγμένες του χρήστη $u$ μετά την προβολή τους στο λανθάνοντα χώρο $k$ διάστασης. Αντίστοιχα, η $i$-οστή γραμμή του $\mathbf{Q}$, $\mathbf{q_i} \in \mathbb{R}^k$, μπορεί να θεωρηθεί ως οι συντεταγμένες του αντικειμένου $i$ σε αυτόν το λανθάνοντα χώρο. Τα μητρώα $\mathbf{P}$ και $\mathbf{Q}$ συνήθως υπολογίζονται ελαχιστοποιώντας το reconstruction error που ορίζεται με τη squared Frobenius norm:

$$\text{err}(\mathbf{P}, \mathbf{Q}) = ||\mathbf{R} - \mathbf{P}\mathbf{Q}^\top||_F^2 = \sum_{u,i} (\mathbf{r_{ui}} - \mathbf{p_u}\mathbf{q_i}^\top)^2.$$

Η ελαχιστοποίηση αυτού του σφάλματος είναι ισοδύναμη με τη διάσπαση σε ιδιοτιμές (*Singular Value Decomposition*, SVD) του μητρώου $\mathbf{R}$ [20]:

$$\mathbf{R} = \mathbf{U\Sigma V}^\top,$$

όπου $\mathbf{U}$ είναι το $|\mathbf{U}| \times n$ μητρώο που περιέχει τα αριστερά ιδιοδιανύσματα, $\mathbf{V}$ είναι το $|\mathbf{I}| \times n$ μητρώο που περιέχει τα δεξιά ιδιοδιανύσματα, και $\mathbf{\Sigma}$ είναι το $n \times n$ διαγώνιο μητρώο με τις ιδιοτιμές. Συμβολίζουμε με $\mathbf{\Sigma_k}, \mathbf{U_k}$ και $\mathbf{V_k}$ τα μητρώα τα οποία προκύπτουν επιλέγοντας ένα υποσύνολο με τις $k$ μεγαλύτερες ιδιοτιμές και τα αντίστοιχα ιδιοδιανύσματα, το παραγοντοποιημένο μητρώο χρηστών και το παραγοντοποιημένο μητρώο αντικειμένων αντιστοιχούν σε $\mathbf{P} = \mathbf{U_k}\mathbf{\Sigma_k}^{1/2}$ και $\mathbf{Q} = \mathbf{V_k}\mathbf{\Sigma_k}^{1/2}$ αντίστοιχα.

Μόλις υπολογιστούν τα $\mathbf{P}$ και $\mathbf{Q}$, η κλασική *model-based* πρόβλεψη μιας βαθμολογίας $r_{ui}$ είναι:

$$r_{ui} = \mathbf{p_u}\mathbf{q_i}^\top$$

Ωστόσο, υπάρχει ένα σημαντικό πρόβλημα στην εφαρμογή του SVD στο μητρώο με τις βαθμολογίες $\mathbf{R}$: οι περισσότερες τιμές $r_{ui}$ του $\mathbf{R}$ δεν έχουν οριστεί, καθώς είναι πολύ πιθανό να μην έχει δοθεί κάποια βαθμολογία στο $i$ από τον $u$. Αν και είναι δυνατό να τεθεί μία καθορισμένη τιμή στο $r_{ui}$, όπως αναφέρθηκε και παραπάνω, κάτι τέτοιο θα εισήγαγε κάποιου είδους προκατάληψη στα δεδομένα. Ακόμα πιο σημαντικό πρόβλημα είναι ότι κάτι τέτοιο θα οδηγούσε στο να κάνει πυκνό το μεγάλο μητρώο $\mathbf{R}$, και κατ' επέκταση να καταστήσει μη πρακτική την SVD διάσπαση του $\mathbf{R}$. Κοινή λύση σε αυτό το πρόβλημα αποτελεί το να μάθουν τα $\mathbf{P}$ και $\mathbf{Q}$ να χρησιμοποιούν μόνο τις γνωστές βαθμολογίες [4, 29, 52, 53]:

$$\text{err}(\mathbf{P}, \mathbf{Q}) = \sum_{r_{ui} \in R} (r_{ui} - \mathbf{p_u}\mathbf{q_i}^\top)^2 + \lambda(||\mathbf{p_u}||^2 + ||\mathbf{q_i}||^2),$$

όπου $\lambda$ είναι μία παράμετρος η οποία ελέγχει το επίπεδο της κανονικοποίησης.

Στις συστάσεις που βασίζονται στη γειτνίαση, μπορεί να χρησιμοποιηθεί η ίδια αρχή για τον υπολογισμό της ομοιότητας ανάμεσα σε χρήστες ή αντικείμενα στο λανθάνοντα χώρο [5]. Αυτό μπορεί να συμβεί λύνοντας το ακόλουθο πρόβλημα:

$$\text{err}(\mathbf{P}, \mathbf{Q}) = \sum_{r_{ui} \in \mathbf{R}} (zui - \mathbf{p_u}\mathbf{q_i}^\top)^2$$

υπό τους περιορισμούς:  $||\mathbf{p_u}|| = 1, \forall u \in \mathbf{U}, ||\mathbf{q_i}|| = 1, \forall i \in \mathbf{I}$,

όπου $z_{ui}$ είναι η βαθμολογία $r_{ui}$ κανονικοποιημένη στο διάστημα $[-1, 1]$. Για παράδειγμα, αν $r_{min}$ και $r_{max}$ είναι η ελάχιστη και μέγιστη τιμή αντίστοιχα στο αρχικό σύνολο βαθμολογιών,

$$z_{ui} = \frac{r_{ui} - \bar{r}_u}{r_{max} - r_{min}}.$$

Αυτό το πρόβλημα αντιστοιχεί στην εύρεση, για κάθε χρήστη $u$ και αντικείμενο $i$, συντεταγμένων στην επιφάνεια της $k$-διάστασης μοναδιαίας σφαίρας έτσι ώστε ο $u$ να δώσει μία υψηλή βαθμολογία στο $i$ αν οι συντεταγμένες τους είναι κοντά μεταξύ τους στην επιφάνεια. Αν δύο χρήστες $u$ και $v$ είναι γειτονικοί στην επιφάνεια, τότε θα δώσουν παρόμοιες βαθμολογίες στα ίδια αντικείμενα, και έτσι, η ομοιότητα ανάμεσα σε αυτούς τους χρήστες μπορεί να υπολογιστεί ως

$$w_{uv} = \mathbf{p_u}\mathbf{p_v}^\top.$$

Αντίστοιχα, η ομοιότητα ανάμεσα σε δύο αντικείμενα $i$ και $j$ μπορεί να υπολογιστεί ως

$$w_{ij} = \mathbf{q_i}\mathbf{q_j}^\top.$$

### 2.2.2   Διάσπαση του μητρώου ομοιοτήτων

Η βασική αρχή αυτής της δεύτερης dimensionality reduction προσέγγισης είναι η ίδια με της προηγούμενης: διάσπαση του μητρώου στους πρωταρχικούς του παράγοντες οι οποίοι αποτελούν την προβολή των χρηστών ή των αντικειμένων στον λανθάνοντα χώρο. Ωστόσο, αντί να διασπαστεί το μητρώο βαθμολογιών, διασπάται ένα αραιό μητρώο ομοιοτήτων.

Έστω $\mathbf{W}$ ένα συμμετρικό μητρώο διάστασης $n$ το οποίο αποτελείται από ομοιότητες είτε χρηστών είτε αντικειμένων. Θα υποθέσουμε την προηγούμενη περίπτωση. Για ακόμα μία φορά, θέλουμε να προσεγγίσουμε το $\mathbf{W}$ με ένα μητρώο $\hat{\mathbf{W}} = \mathbf{P}\mathbf{P}^\top$ μικρότερης διάστασης $k < n$ ελαχιστοποιώντας την ακόλουθη ποσότητα:

$$\text{err}(\mathbf{P}) = ||\mathbf{R} - \mathbf{P}\mathbf{P}^\top||_F^2 = \sum_{u,v} (w_{uv} - \mathbf{p_u}\mathbf{p_v}^\top)^2.$$

Το μητρώο $\hat{\mathbf{W}}$ είναι μία "συμπιεσμένη" έκδοση του $\mathbf{W}$ η οποία είναι λιγότερη αραιή σε σχέση με το $\mathbf{W}$. Όπως και προηγουμένως, η εύρεση του μητρώου παραγόντων $\mathbf{P}$ είναι ισοδύναμη με τον υπολογισμό των ιδιοτιμών του $\mathbf{W}$:

$$\mathbf{W} = \mathbf{V}\mathbf{\Lambda}\mathbf{V}^\top$$

όπου $\mathbf{\Lambda}$ είναι ένα διαγώνιο μητρώο που περιέχει τις $|\mathbf{U}|$ ιδιοτιμές του $\mathbf{W}$, και $\mathbf{V}$ είναι ένα $|\mathbf{U}| \times |\mathbf{U}|$ ορθογώνιο μητρώο που περιέχει τα αντίστοιχα ιδιοδιανύσματα. Έστω $\mathbf{V_k}$ ένα μητρώο που σχηματίζεται από τα $k$ κύρια (κανονικοποιημένα) ιδιοδιανύσματα του $\mathbf{W}$, που αντιστοιχούν στους άξονες του λανθάνοντα υποχώρου διάστασης $k$. Οι συντεταγμένες $\mathbf{p_u} \in \mathbb{R}^k$ ενός χρήστη $u$ σε αυτόν τον υποχώρο δίνεται από την $u$-οστή γραμμή του μητρώου $\mathbf{P} = \mathbf{V_k}\mathbf{\Lambda}_k^{1/2}$. Επιπλέον, οι ομοιότητες χρηστών που υπολογίζονται σε αυτόν το λανθάνοντα υποχώρο δίνονται από το μητρώο

$$\mathbf{W}\prime = \mathbf{P}\mathbf{P}^\top = \mathbf{V_k}\mathbf{\Lambda_k}\mathbf{V_k}^\top$$

## 2.3 Graph-Based Models

Στις graph-based προσεγγίσεις, τα δεδομένα αναπαρίστανται από ένα γράφο όπου οι κόμβοι είναι χρήστες, αντικείμενα ή και τα δύο, και οι ακμές αναπαριστούν τις αλληλεπιδράσεις ή ομοιότητες ανάμεσα στους χρήστες και τα αντικείμενα. Για παράδειγμα, στο Σχήμα 2.3, τα δεδομένα μοντελοποιούνται ως ένα διμερές γράφημα όπου τα δύο σύνολα ακμών αναπαριστούν χρήστες και αντικείμενα, και μία ακμή συνδέει το χρήστη $u$ με το αντικείμενο $i$ αν υπάρχει βαθμολογία στο σύστημα που έχει δοθεί στο $i$ από τον $u$. Στην ακμή μπορεί επίσης να αποδοθεί ένα βάρος, όπως η τιμή της αντίστοιχης βαθμολογίας. Σε ένα άλλο μοντέλο, οι κόμβοι μπορεί να αναπαριστούν είτε χρήστες είτε αντικείμενα, και μία ακμή συνδέει δύο κόμβους αν οι βαθμολογίες που αντιστοιχούν στους δύο αυτούς κόμβους είναι επαρκώς σχετιζόμενες. Το βάρος αυτής της ακμής μπορεί να είναι η αντίστοιχη τιμή συσχέτισης [17].

Σε αυτά τα μοντέλα, οι στάνταρ προσεγγίσεις που βασίζονται στις συσχετίσεις προβλέπουν τη βαθμολογία ενός χρήστη $u$ για ένα αντικείμενο $i$ χρησιμοποιώντας μόνο τους κόμβους οι οποίοι συνδέονται απευθείας με τον $u$ ή το $i$. Από την άλλη μεριά, οι graph-based προσεγγίσεις επιτρέπουν σε κόμβους οι οποίοι δε συνδέονται απευθείας να επηρεάζουν ο ένας τον άλλο διαδίδοντας πληροφορίες κατά μήκος των ακμών του γραφήματος. Όσο μεγαλύτερο είναι το βάρος μιας ακμής, τόσο περισσότερες πληροφορίες επιτρέπεται να περάσουν δια μέσω αυτής. Επίσης, η επιρροή ενός κόμβου σε έναν άλλο πρέπει να είναι μικρότερη αν οι δύο κόμβοι είναι πολύ μακριά στο γράφημα. Αυτές οι δύο ιδιότητες, γνωστές ως propagation και attenuation [23, 24], παρατηρούνται συχνά σε graph-based μέτρα ομοιότητας.

Οι μεταβατικές συσχετίσεις που συλλαμβάνονται από τις graph-based μεθόδους μπορούν να χρησιμοποιηθούν για τη σύσταση αντικειμένων με δύο διαφορετικούς τρόπους. Στην πρώτη

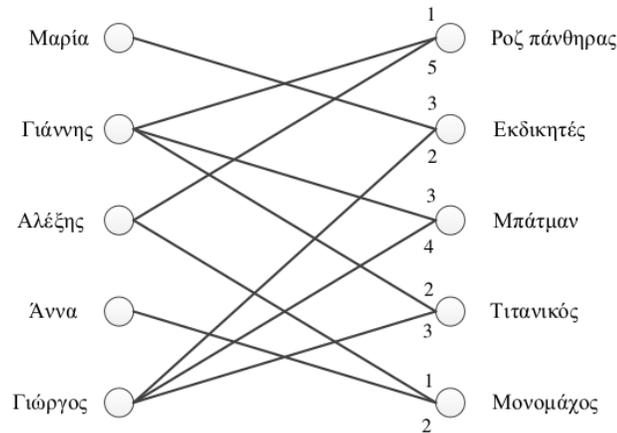

ΣΧΉΜΑ 2.3: Διμερής γράφος ο οποίος παρουσιάζει τις βαθμολογίες των χρηστών από τον Πίνακα 2.1. Οι βαθμολογίες αντιστοιχούν σε βάρη των ακμών. [32].

προσέγγιση, η εγγύτητα ενός χρήστη $u$ σε ένα αντικείμενο $i$ στο γράφημα χρησιμοποιείται απευθείας για την αξιολόγηση της βαθμολογίας του $u$ στο $i$ [18, 23, 24]. Ακολουθώντας αυτήν την ιδέα, τα αντικείμενα που συστήνονται από το σύστημα στον $u$ είναι εκείνα τα οποία είναι τα πιο "κοντινά" στον $u$ στο γράφημα. Από την άλλη μεριά, η δεύτερη προσέγγιση θεωρεί την εγγύτητα δύο κόμβων χρηστών ή αντικειμένων στο γράφο ως μέτρο ομοιότητας, και χρησιμοποιεί αυτήν την ομοιότητα ως τα βάρη μια μεθόδου συστάσεων που βασίζεται στη γειτονικότητα [18, 35].

### 2.3.1 Path-based ομοιότητα

Σε αυτές τις μεθόδους, η απόσταση ανάμεσα σε δύο κόμβους του γράφου εκτιμάται ως συνάρτηση του πλήθους των μονοπατιών που συνδέουν τους δύο κόμβους, καθώς και το μήκος αυτών των μονοπατιών.

**Συντομότερο μονοπάτι.** Στο [1] περιγράφεται μία μέθοδος συστάσεων η οποία υπολογίζει την ομοιότητα ανάμεσα σε δύο χρήστες με βάση τη μικρότερη απόσταση μεταξύ τους σε ένα γράφο. Σε αυτήν τη μέθοδο, τα δεδομένα μοντελοποιούνται ως ένας κατευθυνόμενος γράφος του οποίου οι κόμβοι είναι χρήστες και οι ακμές καθορίζονται με βάση τις έννοιες *horting* και *predictability*. Με την έννοια horting εννοούμε μία σχέση ανάμεσα σε δύο χρήστες η οποία ικανοποιείται αν αυτοί οι χρήστες έχουν βαθμολογήσει παρόμοια αντικείμενα. Από την άλλη μεριά, με την έννοια predictability αναφερόμαστε σε μία πιο ισχυρή ιδιότητα η οποία απαιτεί επιπλέον οι βαθμολογίες του ενός χρήστη να είναι παρόμοιες με αυτές του άλλου.

**Σύνολο μονοπατιών.** Εναλλακτικά, για να εκτιμηθεί η συμβατότητα ενός χρήστη και ενός αντικειμένου σε ένα διμερή γράφο, μπορεί να χρησιμοποιηθεί ο αριθμός των μονοπατιών ανάμεσά τους [24]. Έστω $\mathbf{R}$ το $|\mathbf{U}| \times |\mathbf{I}|$ μητρώο βαθμολογιών όπου $r_{ui}$ ισούται με 1 αν ο χρήστης $u$ έχει βαθμολογήσει το αντικείμενο $i$, και $0$ αλλιώς. Το μητρώο γειτνίασης $\mathbf{A}$ του διμερούς γράφου μπορεί να οριστεί από το $\mathbf{R}$ ως

$$\mathbf{A} = \begin{pmatrix} \mathbf{0} & \mathbf{R}^\top \\ \mathbf{R} & \mathbf{0} \end{pmatrix}.$$

Σε αυτήν την προσέγγιση, η σχέση ανάμεσα σε ένα χρήστη $u$ και ένα αντικείμενο $i$ ορίζεται ως το άθροισμα των βαρών όλων των διακριτών μονοπατιών που συνδέουν τον $u$ με τον $v$, όπου $v$ ένας δεύτερος χρήστης (επιτρέποντας στους κόμβους να εμφανίζονται περισσότερες από μία φορές στο μονοπάτι), των οποίων το μήκος δεν είναι μεγαλύτερο από ένα μέγιστο μήκος $K$. Να σημειωθεί ότι από τη στιγμή που ο γράφος είναι διμερής, το $K$ πρέπει να είναι περιττός αριθμός. Προκειμένου να ελαττωθεί η συμβολή των μεγαλύτερων μονοπατιών, το βάρος που δίνεται σε ένα μονοπάτι μήκους $k$ ορίζεται ως $\alpha^k$, όπου $\alpha \in [0,1]$. Με βάση το γεγονός ότι ο αριθμός των μονοπατιών μήκους $k$ μεταξύ ζευγαριών από κόμβους δίνεται από το $\mathbf{A^k}$, το μητρώο συσχετίσεων χρηστών-αντικειμένων $\mathbf{S_k}$ είναι

$$\mathbf{S_K} = \sum_{k=1}^{K} \alpha^k \mathbf{A^k} = (I - \alpha \mathbf{A})^{-1}(\alpha \mathbf{A} - \alpha^K \mathbf{A^K}).$$

Αυτή η μέθοδος υπολογισμού αποστάσεων ανάμεσα σε κόμβους ενός γράφου είναι γνωστή ως μέτρο *Katz* [26].

Στα συστήματα συστάσεων τα οποία έχουν μεγάλο αριθμό χρηστών και αντικειμένων, ο υπολογισμός της παραπάνω τιμής συσχέτισης, αλλά και άλλων αντίστοιχων τιμών, ίσως απαιτεί εκτεταμένους υπολογιστικούς πόρους. Για την αντιμετώπιση αυτών των περιορισμών, στο [24] χρησιμοποιήθηκαν τεχνικές διάδοσης ενεργοποίησης (spreading activation techniques) [14]. Ουσιαστικά, τέτοιες τεχνικές λειτουργούν ως εξής: αρχικά ενεργοποιούν ένα επιλεγμένο υποσύνολο κόμβων ως κόμβους εκκίνησης, και στη συνέχεια επαναληπτικά ενεργοποιούν τους κόμβους που προσπελάζονται απευθείας από τους κόμβους που είναι ήδη ενεργοί, μέχρι να υπάρξει κάποιο κριτήριο σύγκλισης.

### 2.3.2 Random walk ομοιότητα

Μία δεύτερη μέθοδος υπολογισμού της ομοιότητας είναι η *random walk* ομοιότητα, όπου οι μεταβατικές σχέσεις ορίζονται μέσα σε ένα πιθανοτικό πλαίσιο. Σε αυτό το πλαίσιο, η ομοιότητα ή συνάφεια ανάμεσα σε χρήστες ή αντικείμενα εκτιμάται ως η πιθανότητα προσπέλασης αυτών των κόμβων σε έναν τυχαίο περίπατο. Τυπικά, αυτό μπορεί να περιγραφεί με μία πρώτης τάξης Μαρκοβιανή διαδικασία με μητρώο πιθανοτήτων μετάβασης $\mathbf{P} \in \mathbb{R}^{n \times n}$ [17]. Η πιθανότητα μετάβασης από την κατάσταση $i$ στη $j$ σε οποιοδήποτε χρονικό βήμα $t$ είναι

$$p_{ij} = \Pr(s(t+1) = j | s(t) = i).$$

Έστω $\boldsymbol{\pi}(t)$ το διάνυσμα το οποίο περιέχει την κατανομή καταστάσεων για το βήμα $t$, έτσι ώστε $\pi_i(t) = \Pr(s(t) = i)$. Η εξέλιξη της Μαρκοβιανής αλυσίδας χαρακτηρίζεται από

$$\boldsymbol{\pi}(t+1) = \mathbf{P}^\top \boldsymbol{\pi}(t).$$

Επιπλέον, υπό την προϋπόθεση ότι το $\mathbf{P}$ είναι στοχαστικό κατά γραμμές, δηλαδή $\sum_j p_{ij}$ για όλα τα $i$, η διαδικασία συγκλίνει σε ένα διάνυσμα σταθερής κατανομής $\boldsymbol{\pi}(\infty)$ που αντιστοιχεί στο θετικό ιδιοδιάνυσμα του $\mathbf{P}^\top$ με ιδιοτιμή 1. Αυτή η διαδικασία περιγράφεται συχνά με τη μορφή ενός γράφου με βάρη ο οποίος έχει έναν κόμβο για κάθε κατάσταση, και όπου η πιθανότητα να υπάρξει μετάβαση από έναν κόμβο σε ένα διπλανό κόμβο δίνεται από το βάρος της ακμής που συνδέει αυτούς τους κόμβους.

**ItemRank.** Μία μέθοδος συστάσεων η οποία βασίζεται στον αλγόριθμο PageRank για την ταξινόμηση ιστοσελίδων [9], είναι ο ItemRank [23]. Αυτή η προσέγγιση ταξινομεί τις προτιμήσεις ενός χρήστη $u$ για νέα αντικείμενα $i$ με βάση την πιθανότητα ο $u$ να επισκεφτεί το $i$ κατά τη διάρκεια ενός τυχαίου περιπάτου σε ένα γράφο όπου οι κόμβοι αντιστοιχούν στα αντικείμενα του συστήματος και οι ακμές συνδέουν αντικείμενα που έχουν βαθμολογηθεί από κοινούς χρήστες. Τα βάρη των ακμών δίνονται από το $|\mathbf{I}| \times |\mathbf{I}|$ μητρώο πιθανοτήτων μετάβασης $\mathbf{P}$ για το οποίο $p_{ij} = |U_{ij}|/|U_i|$ είναι η αναμενόμενη δεσμευμένη πιθανότητα ένας χρήστης να βαθμολογήσει ένα αντικείμενο $j$ αν έχει βαθμολογήσει ένα αντικείμενο $i$.

Όπως και στον PageRank, ο τυχαίος περιηγητής μπορεί σε οποιοδήποτε βήμα $t$, είτε χρησιμοποιώντας το $\mathbf{P}$ να μεταπηδήσει σε ένα γειτονικό κόμβο με πιθανότητα $\alpha$, είτε να "τηλεμεταφερθεί" σε οποιοδήποτε κόμβο με πιθανότητα $(1-\alpha)$. Έστω $\mathbf{r}_u$ η $u$-οστή γραμμή του μητρώου με τις βαθμολογίες $\mathbf{R}$. Τότε η κατανομή πιθανότητας του χρήστη $u$ να τηλεμεταφερθεί σε άλλους κόμβους δίνεται από το διάνυσμα $\mathbf{d_u} = \mathbf{r_u}/||\mathbf{r_u}||$. Με βάση αυτούς τους ορισμούς, το διάνυσμα σταθερής κατανομής του χρήστη $u$ κατά το βήμα $t+1$ μπορεί να εκφραστεί αναδρομικά ως

$$\boldsymbol{\pi_u}(t+1) = \alpha \mathbf{P}^\top \boldsymbol{\pi_u}(t) + (1-\alpha)\mathbf{d_u}. \tag{2.1}$$

Για πρακτικούς λόγους, το $\boldsymbol{\pi_u}(\infty)$ συνήθως υπολογίζεται από μία διαδικασία η οποία πρώτα αρχικοποιεί την κατανομή ως ομοιόμορφη, δηλαδή $\boldsymbol{\pi_u}(0) = \dfrac{1}{n}\mathbf{1}_n$, και στη συνέχεια επαναληπτικά ενημερώνει το $\boldsymbol{\pi_u}$, χρησιμοποιώντας την 2.1, έως ότου συγκλίνει. Μόλις το $\boldsymbol{\pi_u}(\infty)$ έχει υπολογιστεί, το σύστημα συστήνει στον $u$ το αντικείμενο $i$ το οποίο έχει το υψηλότερο $\pi_{ui}$.

## 2.4   Συμπεράσματα

Παρότι εφαρμόζονται με επιτυχία σε πολλές εφαρμογές, οι στάνταρ CF τεχνικές αντιμετωπίζουν πολλές προκλήσεις οι οποίες δεν έχουν επιλυθεί ακόμα. Μία από τις σημαντικότερες, όπως αναφέρθηκε και παραπάνω, είναι η *αραιότητα*, ένα πολύ σύνηθες πρόβλημα το οποίο προκύπτει όταν τα διαθέσιμα δεδομένα δεν επαρκούν για την αναγνώριση παρόμοιων στοιχείων [8, 38, 40, 43, 44]. Η αραιότητα είναι ένα εγγενές χαρακτηριστικό των συστημάτων συστάσεων διότι στη μεγαλύτερη πλειοψηφία των πραγματικών εφαρμογών οι χρήστες αλληλεπιδρούν μόνο με ένα μικρό ποσοστό των διαθέσιμων αντικειμένων, και την ίδια στιγμή νέοι χρήστες και νέα αντικείμενα προστίθενται τακτικά στο σύστημα. Οι παραδοσιακές CF τεχνικές, όπως τα neighborhood models, παρουσιάζουν ευπάθεια στην αραιότητα, ένα γεγονός το οποίο περιορίζει την ποιότητα των συστάσεων που παράγουν [17].

Ανάμεσα στις πιο υποσχόμενες προσεγγίσεις για την αντιμετώπιση των προβλημάτων που σχετίζονται με αυτό το χαρακτηριστικό είναι τα *Latent Factor* και *Graph-Based* μοντέλα. Ωστόσο, παρότι οι τεχνικές αυτές είναι πολλά υποσχόμενες όσον αφορά στο χειρισμό των προβλημάτων που σχετίζονται με την αραιότητα, αντιμετωπίζουν σοβαρά υπολογιστικά θέματα και περιορισμούς επεκτασιμότητας, καθώς ο αριθμός των χρηστών και των αντικειμένων αυξάνεται ραγδαία στις σύγχρονες εφαρμογές ηλεκτρονικού εμπορίου.

Οι Nikolakopoulos et al. [38, 40, 43, 44] για την αντιμετώπιση της αραιότητας και των προβλημάτων που προκαλεί η ύπαρξή της, εκμεταλλεύονται την ιεραρχική δομή του χώρου αντικειμένων [37] και τα μεταδεδομένα από τα datasets, και προτείνουν προσεγγίσεις οι οποίες τα πηγαίνουν ιδιαίτερα καλά, ενώ ταυτόχρονα παραμένουν υπολογιστικά αποδοτικές και επεκτάσιμες. Πιο συγκεκριμένα, οι συγγραφείς στο [44] πρότειναν τον αλγόριθμο *Hierarchical Itemspace Rank* (HIR), ο οποίος ανακαλύπτει σχέσεις ανάμεσα στα αντικείμενα μέσα από τη διάσπαση του χώρου αντικειμένων. Ορίζει μπλοκ από στενά σχετιζόμενα στοιχεία και χρησιμοποιώντας αυτή τη νέα σύνθεση εκμεταλλεύεται τις ιδιότητες που είναι κρυμμένες στη δομή του χώρου αντικειμένων.

Επιπρόσθετα, οι συγγραφείς προτείνουν [38, 40] ένα γενικό τρόπο αντιμετώπισης της αραιότητας και των συνεπειών της μέσω της μεθόδου *NCDREC*. Πρόκειται για μία αποτελεσματική και επεκτάσιμη προσέγγιση η οποία συνδυάζει την αποτελεσματικότητα των latent factor μοντέλων με την ικανότητα των graph-based μοντέλων να διατηρούν τις "τοπικές" σχέσεις ανάμεσα στα στοιχεία. Και πάλι βασιζόμενοι στην ιδέα της *Decomposability* [37], μέσα από τη διάσπαση του χώρου αντικειμένων ορίζουν μπλοκ από στενά σχετιζόμενα στοιχεία και εισάγουν τα αντίστοιχα έμμεσα συστατικά εγγύτητας τα οποία έχουν ως στόχο να γεμίσουν τα κενά στα δεδομένα που προκαλεί η αραιότητα. Τέλος, μελετούν τις θεωρητικές ιδιότητες αυτής της διάσπασης και παρουσιάζουν τις επαρκείς συνθήκες που εγγυώνται πλήρη κάλυψη του χώρου αντικειμένων ακόμα και σε καταστάσεις έντονης αραιότητας.

Ο στόχος μας σε αυτήν την εργασία είναι να βρεθεί μία εναλλακτική μέθοδος συστάσεων η οποία να συνδυάζει υπολογιστική αποδοτικότητα και να μην παρουσιάζει ευαισθησία στην αραιότητα, χωρίς ταυτόχρονα να θυσιάζει την top-N ποιότητα. Ακολουθώντας την προσέγγιση του PureSVD, επικεντρωνόμαστε κι εμείς στη μείωση της διάστασης του προβλήματος, αλλά με πιο συμφέρων τρόπο. Έτσι, προτείνουμε τη δημιουργία ενός latent factor μοντέλου εκμεταλλευόμενοι μία υπολογιστικά αποδοτική διαδικασία Krylov υποχώρου, η οποία ονομάζεται *Lanczos method*.

# Κεφάλαιο 3

# Lanczos Latent Factor Recommender

## 3.1 Latent Factor Models

Η βασική υπόθεση πίσω από τη χρήση των latent factor μοντέλων για τη δημιουργία συστημάτων συστάσεων είναι ότι οι προτιμήσεις των χρηστών επηρεάζονται από ένα σύνολο "κρυμμένων παραγόντων προτίμησης" οι οποίοι είναι συνήθως πολύ συγκεκριμένοι στον τομέα των συστάσεων [30]. Αυτοί οι παράγοντες είναι γενικά μη εμφανείς και μπορεί να μην είναι απαραίτητα διαισθητικά κατανοητοί. Ωστόσο, οι Latent Factor αλγόριθμοι μπορούν να συμπεράνουν αυτούς τους παράγοντες από την ανατροφοδότηση του χρήστη όπως αυτή αντικατοπτρίζεται στις βαθμολογίες.

### 3.1.1 PureSVD

Ένα από τα πιο επιτυχημένα παραδείγματα latent factor αλγορίθμων για top-N συστάσεις, είναι ο PureSVD. Ο αλγόριθμος PureSVD θεωρεί μηδενικές όλες τις τιμές που λείπουν από το μητρώο με τις βαθμολογίες, και παράγει συστάσεις προσεγγίζοντας το user-item rating μητρώο $\mathbf{R}$ με την παρακάτω παραγοντοποίηση:

$$\hat{\mathbf{R}} = \mathbf{U}\mathbf{\Sigma}\mathbf{Q}^\top$$

όπου, $\mathbf{U}$ είναι ένα $n \times f$ ορθοκανονικό μητρώο, $\mathbf{Q}$ είναι ένα $m \times f$ ορθοκανονικό μητρώο, και $\mathbf{\Sigma}$ είναι ένα $f \times f$ διαγώνιο μητρώο που περιέχει τις πρώτες $f$ ιδιάζουσες τιμές. Οι γραμμές του μητρώου $\hat{\mathbf{R}}$ περιλαμβάνουν τα διανύσματα συστάσεων για κάθε χρήστη στο σύστημα. Να σημειωθεί ότι παρότι οι πραγματικές τιμές του μητρώου $\hat{\mathbf{R}}$ δεν έχουν νόημα ως βαθμολογίες, εισάγουν μία διάταξη στα αντικείμενα η οποία είναι αρκετή για τη σύσταση top-N λιστών.

Οι συγγραφείς στο [13], μετά την αξιολόγηση της απόδοσης διαφόρων latent factor-based αλγορίθμων και neighborhood μοντέλων, βρήκαν ότι ο PureSVD ήταν ικανός να παράγει καλύτερες top-N συστάσεις συγκριτικά με εξελιγμένες matrix factorization μεθόδους [29, 31] και άλλες διάσημες CF τεχνικές. Ωστόσο, παρά τα πολύ καλά του αποτελέσματα, ο PureSVD περιλαμβάνει τον υπολογισμό του truncated singular value decomposition του μητρώου με τις βαθμολογίες, ο οποίος, λόγω του αυξανόμενου αριθμού χρηστών και αντικειμένων στις μοντέρνες εφαρμογές ηλεκτρονικού εμπορίου, θα μπορούσε να επιφέρει απαγορευτικά μεγάλο υπολογιστικό βάρος.

## 3.2 Το μοντέλο LLFR

**Μέθοδος Lanczos.** Η μέθοδος Lanczos χρησιμοποιήθηκε αρχικά σε εφαρμογές γραμμικής άλγεβρας για τον υπολογισμό των ιδιοδιανυσμάτων ή/και των ιδιαζουσών τριπλετών μεγάλων αραιών μητρώων [21]. Από ποιοτική άποψη, οι Blom και Ruhe [7] πρότειναν τη χρήση ενός αλγορίθμου στενά συνδεδεμένου με τη μέθοδο Latent Semantic Indexing ο οποίος χρησιμοποιεί τη Lanczos τεχνική διδιαγωνοποίησης για να παράξει δύο σύνολα διανυσμάτων τα οποία ουσιαστικά αντικαθιστούν τα αριστερά και δεξιά ιδιάζοντα διανύσματα, μειώνοντας το υπολογιστικό κόστος.

Οι Chen and Saad [12] εξέτασαν πρόσφατα τη χρήση των Lanczos διανυσμάτων σε εφαρμογές όπου το κύριο ζήτημα μετατίθεται στον υπολογισμό ενός γινομένου μητρώου-διανύσματος στις κύριες ιδιάζουσες κατευθύνσεις του μητρώου δεδομένων. Έδειξαν την αποτελεσματικότητα αυτής της προσέγγισης σε δύο διαφορετικά προβλήματα από το χώρο της ανάκτησης πληροφορίας και αναγνώρισης προσώπου.

Απ' όσο είμαστε σε θέση να γνωρίζουμε, πρόκειται για την πρώτη εργασία η οποία προτείνει τη χρήση των Lanczos διανυσμάτων για το πρόβλημα των top-N συστάσεων. Επιπλέον, ο στόχος μας ήταν διαφορετικός από τους παραπάνω, υπό την έννοια ότι εμείς εφαρμόζουμε τη μέθοδο Lanczos απευθείας σε ένα μητρώο ομοιότητας αντικειμένων χωρίς να προσπαθήσουμε να ακολουθήσουμε την προσέγγιση του PureSVD.

**Ορισμοί.** Έστω $\mathcal{U} = \{u_1, u_2, \ldots, u_n\}$ ένα σύνολο από *χρήστες* και $\mathcal{V} = \{v_1, v_2, \ldots, v_m\}$ ένα σύνολο από *αντικείμενα*. Έστω $\mathcal{R}$ ένα σύνολο από πλειάδες $t_{ij} = (u_i, v_j, r_{ij})$, όπου $r_{ij}$ είναι ένας μη αρνητικός αριθμός στον οποίο θα αναφερόμαστε ως η *βαθμολογία* που δίνεται από το χρήστη $u_i$ για το αντικείμενο $v_j$, και έστω $\mathbf{R} \in \Re^{n \times m}$ ένα μητρώο του οποίου το $ij^{th}$ στοιχείο περιέχει τη βαθμολογία $r_{ij}$ αν η πλειάδα $t_{ij}$ ανήκει στο $\mathcal{R}$, και μηδέν αλλιώς. Αυτές οι βαθμολογίες μπορούν να προέρχονται είτε από την ξεκάθαρη απόκριση του χρήστη είτε να συμπεραίνονται από τη συμπεριφορά και αλληλεπίδραση του χρήστη με το σύστημα. Επίσης, θεωρούμε μια διαμέριση $\{\mathcal{L}, \mathcal{T}\}$ των βαθμολογιών σε ένα *σύνολο εκπαίδευσης - training set* $\mathcal{L}$ και ένα *σύνολο ελέγχου - test set* $\mathcal{T}$. Για κάθε χρήστη $u_i$, δηλώνουμε ως $\mathcal{L}_i$ το σύνολο των αντικειμένων που έχουν βαθμολογηθεί από τον $u_i$ στο $\mathcal{L}$. Πιο συγκεκριμένα:

$$\mathcal{L}_i \triangleq \{v_k : t_{ik} \in \mathcal{L}\}$$

**Μητρώο Συσχετίσεων μεταξύ Αντικειμένων (Inter-item Correlation Matrix) A**. Αρχικά είναι απαραίτητο να ορίσουμε ένα μητρώο το οποίο να συλλαμβάνει τις ομοιότητες ανάμεσα στα στοιχεία του χώρου αντικειμένων. Έτσι, ορίζουμε ένα συμμετρικό μητρώο $\mathbf{A} \in \mathfrak{R}^{m \times m}$ του οποίου το στοιχείο $ij^{th}$ δίνεται από:

$$A_{k\ell} \triangleq \|\mathbf{r_k}\| \|\mathbf{r_\ell}\| |\mathcal{U}_{k\ell}|, \tag{3.1}$$

όπου $\|\mathbf{r_j}\|$ είναι η ευκλείδεια νόρμα της στήλης που αντιστοιχεί στο αντικείμενο $v_j$ του μητρώου βαθμολογιών, και το $\mathcal{U}_{k\ell} \subseteq \mathcal{U}$ υποδηλώνει το σύνολο των χρηστών που έχουν βαθμολογήσει και τα δύο αντικείμενα $v_k$ και $v_\ell$, δηλαδή:

$$\mathcal{U}_{k\ell} \triangleq \begin{cases} \{u_s : (v_k \in \mathcal{L}_s) \land (v_\ell \in \mathcal{L}_s)\} & \text{για } k \neq \ell \\ \emptyset & \text{αλλιώς} \end{cases} \tag{3.2}$$

Ο στόχος μας είναι να διασπάσουμε το μητρώο στους κύριους παράγοντές του οι οποίοι αναπαριστούν προβολές των διανυσμάτων συσχέτισης μεταξύ των αντικειμένων στο λανθάνοντα χώρο.

**Κατασκευάζοντας το Λανθάνον Μοντέλο.** Δοθέντος ενός μητρώου $\mathbf{X}$ και ενός αρχικού μοναδιαίου διανύσματος $\mathbf{q}$, ο αντίστοιχος Krylov υποχώρος δίνεται από

$$\mathcal{K}_f(\mathbf{X}, \mathbf{q}) = \text{range}\{\mathbf{q}, \mathbf{Xq}, \mathbf{X^2q}, \ldots, \mathbf{X}^{f-1}\mathbf{q}\} \tag{3.3}$$

Οι μέθοδοι υποχώρου Krylov σχηματίζοντας μία ορθογώνια βάση για τον $\mathcal{K}_f$, μπορούν να χρησιμοποιηθούν για την επίλυση διαφόρων ειδών αριθμητικών προβλημάτων [21]. Η γενική λύση για την εύρεση των ιδιοτιμών/ιδιοδιανυσμάτων που χρησιμοποιεί τον $\mathcal{K}_f$ ονομάζεται μέθοδος Arnoldi και χρησιμοποιεί αναδρομικά όλες τις διαστάσεις του $\mathcal{K}_f$ σε κάθε επανάληψη. Η συμμετρική εκδοχή της Arnoldi ονομάζεται μέθοδος Lanczos και το βασικό της χαρακτηριστικό είναι ότι εκμεταλλεύεται τη συμμετρία του αρχικού μητρώου και χρησιμοποιεί αναδρομή τριών όρων. Κάτι τέτοιο καθιστά τη Lanczos ιδιαίτερα αρμοστή για εφαρμογές σε μεγάλα σύνολα δεδομένων (για περαιτέρω πληροφορίες βλ. [21]).

## 3.3 Ο Αλγόριθμος LLFR

Οι συγκεκριμένες ιδιότητες του μοντέλου μας (συμμετρία και αραιότητα), και το γεγονός ότι ενδιαφερόμαστε για συστάσεις που βασίζονται στην κατάταξη, η οποία μας δίνει την ευελιξία να μη νοιαζόμαστε για τις ακριβείς βαθμολογίες κατάταξης (είναι αρκετή η σωστή σειρά αντικειμένων), καθιστά την προσέγγιση Lanczos έναν ιδανικό υποψήφιο για την κατασκευή του λανθάνοντα χώρου και την αποδοτική παραγωγή λιστών συστάσεων.

Επίσημα, για κάθε χρήστη $u_i$ ορίζουμε ένα προσωποποιημένο διάνυσμα συστάσεων το οποίο δίνεται από:

$$\boldsymbol{\pi_i}^\top \triangleq \mathbf{r_i}^\top \mathbf{Q}\mathbf{Q}^\top \tag{3.4}$$

όπου $\mathbf{r_i}^\top$ οι βαθμολογίες του χρήστη $u_i$ και $\mathbf{Q} \in \mathfrak{R}^{m \times f}$ είναι το μητρώο το οποίο περιλαμβάνει τα Lanczos διανύσματα που σχηματίζουν τη βάση του Krylov υποχώρου $\mathcal{K}_f$ που αντιστοιχεί στο μητρώο συσχετίσεων μεταξύ αντικειμένων $\mathbf{A}$. Ο πλήρης αλγόριθμος για τον υπολογισμό του μητρώου $\mathbf{Q}$ και το τελικό μητρώο συστάσεων $\mathbf{\Pi}$ για όλο το σύνολο των χρηστών δίνεται παρακάτω:

---
**Algorithm 1** Lanczos Latent Factor Recommender (LLFR)

---
**Είσοδος:** Το Μητρώο Συσχετίσεων μεταξύ αντικειμένων $\mathbf{A} \in \mathfrak{R}^{m \times m}$, το Μητρώο Βαθμολογιών $\mathbf{R} \in \mathfrak{R}^{n \times m}$, ένα τυχαίο μοναδιαίο διάνυσμα $\mathbf{q_1} \in \mathfrak{R}^m$, και ο αριθμός των latent factors $f$.
**Έξοδος:** Το μητρώο $\mathbf{\Pi} \in \mathfrak{R}^{n \times m}$ του οποίου οι γραμμές είναι τα διανύσματα συστάσεων για κάθε χρήστη.

1: $\mathbf{q_0} \leftarrow \mathbf{0}$
2: $\beta_1 \leftarrow 0$
3: **for** $i \leftarrow 1, ..., f$ **do**
4: $\quad \mathbf{w} \leftarrow \mathbf{A}\mathbf{q_i} - \beta_i \mathbf{q_{i-1}}$
5: $\quad \alpha_i \leftarrow \mathbf{w}^\top \mathbf{q_i}$
6: $\quad \mathbf{w} \leftarrow \mathbf{w} - \alpha_i \mathbf{q_i}$
7: $\quad \beta_{i+1} \leftarrow \|\mathbf{w}\|_2$
8: $\quad \mathbf{q_{i+1}} \leftarrow \mathbf{w}/\beta_{i+1}$
9: **end for**
10: **return** $\mathbf{\Pi} \leftarrow \mathbf{R}\mathbf{Q}\mathbf{Q}^\top$

---

### 3.3.1 Υπολογιστικά Θέματα

Ο αλγόριθμος LLFR είναι κατά πολύ πιο οικονομικός και σε χρήση μνήμης και σε κατανάλωση χρόνου συγκριτικά με τον PureSVD. Πιο συγκεκριμένα, ο LLFR χρειάζεται $\mathcal{O}((nnz + m)f)$ χρόνο για αραιά μητρώα (όπου $nnz$ είναι ο αριθμός των μη μηδενικών στοιχείων του $\mathbf{A}$), ο οποίος είναι ο χρόνος για τον υπολογισμό των Lanczos διανυσμάτων [12, 21]. Να σημειωθεί ότι σε πραγματικές εφαρμογές, δε χρειάζεται ο εκ των προτέρων υπολογισμός και η αποθήκευση του μητρώου $\mathbf{\Pi}$. Οι αντίστοιχες γραμμές του μπορούν να ανακατασκευαστούν στη στιγμή, όποτε χρειάζεται, από το μικρής διάστασης μητρώο $\mathbf{Q}$.

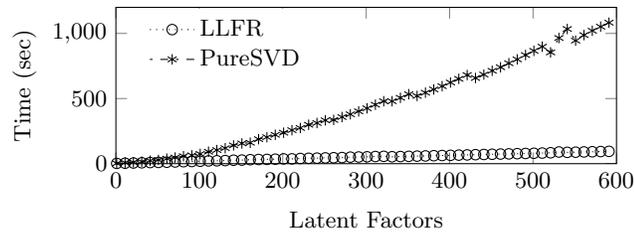

ΣΧΉΜΑ 3.1: Υπολογιστικοί έλεγχοι

Το Σχήμα 3.1 παρουσιάζει το χρόνο υπολογισμού ως συνάρτηση του αριθμού των latent factors για τους αλγόριθμους LLFR και PureSVD όταν εφαρμόστηκαν στο `MovieLens10M` σύνολο δεδομένων[1]. Για τον υπολογισμό του PureSVD χρησιμοποιήθηκε η βελτιστοποιημένη sparse svd συνάρτηση `svds` της `Matlab`. Η διαδικασία LLFR υλοποιήθηκε επίσης σε `Matlab` για να υπάρχει δικαιοσύνη στις συγκρίσεις. Όλα τα πειράματα εκτελέστηκαν σε Linux σύστημα με 64bit μηχανή και 20GB RAM. Το σχήμα 3.1 κάνει φανερό ότι το υπολογιστικό βάρος του LLFR είναι σημαντικά μικρότερο από αυτό του PureSVD. Επιπρόσθετα, όπως θα δούμε στο επόμενο κεφάλαιο, αυτό το πλεονέκτημα υφίσταται χωρίς να θυσιαστεί η ποιότητα των αποτελεσμάτων.

ΠΊΝΑΚΑΣ 3.1: Σύνολα Δεδομένων

| Σύνολο Δεδομένων | #Χρήστες | #Αντικείμενα | #Βαθμολογίες |
|---|---:|---:|---:|
| `MovieLens10M`[2] | 71,567 | 10,681 | 10,000,054 |
| `Yahoo!Music`[3] | 1,823,179 | 136,736 | 717,872,016 |

---

[1] Περισσότερες πληροφορίες για τα σύνολα δεδομένων που χρησιμοποιήθηκαν σε αυτήν την εργασία παρουσιάζονται στον Πίνακα 3.1.
[2] http://grouplens.org/
[3] http://webscope.sandbox.yahoo.com

# Κεφάλαιο 4

# Πειραματική Αξιολόγηση

## 4.1 Datasets

Τα σύνολα δεδομένων τα οποία χρησιμοποιήσαμε για τη διεξαγωγή των πειραμάτων μας είναι τα `MovieLens10M` και `Yahoo!Music`.

Η ερευνητική ομάδα GroupLens συγκέντρωσε και διαθέτει δεδομένα από το διαδικτυακό τόπο MovieLens [1], μία online υπηρεσία συστάσεων ταινιών. Το `MovieLens10M` που χρησιμοποιήσαμε, όπως φαίνεται και στον Πίνακα 3.1, περιέχει $71,567$ χρήστες, $10,681$ αντικείμενα και συνολικά $10,000,054$ βαθμολογίες. Οι χρήστες που συμπεριλήφθηκαν επιλέχθηκαν με τυχαίο τρόπο, και όλοι έχουν βαθμολογήσει τουλάχιστον 20 ταινίες. Στο συγκεκριμένο MovieLens σύνολο δεδομένων δεν περιλαμβάνεται οποιαδήποτε δημογραφική πληροφορία. Τέλος, κάθε χρήστης αναπαρίσταται από ένα id, και δεν παρέχεται καμία περεταίρω πληροφορία.

Αντίστοιχα, το `R2 - Yahoo! Music` αποτελείται από $717,872,016$ βαθμολογίες για $136,736$ αντικείμενα που δόθηκαν από $1,823,179$ χρήστες των υπηρεσιών Yahoo! Music. Το Yahoo! Music αποτελεί ένα στιγμιότυπο των προτιμήσεων της κοινότητας Yahoo! Music για διάφορα τραγούδια. Τα δεδομένα συλλέχθηκαν ανάμεσα στο 2002 και 2006. Κάθε τραγούδι του συνόλου συνοδεύεται από τα εξής χαρακτηριστικά: καλλιτέχνη, άλμπουμ, και κατηγορία. Οι χρήστες, τα τραγούδια, οι καλλιτέχνες και τα άλμπουμ αναπαρίστανται από τυχαία αριθμητικά αναγνωριστικά που τους έχουν αποδοθεί ώστε να μην αποκαλύπτεται καμία αναγνωριστική πληροφορία.

Τα παραπάνω σύνολα δεδομένων χρειάστηκε να τα επεξεργαστούμε κατάλληλα προκειμένου να χρησιμοποιηθούν αποδοτικά στα πειράματά μας.

---

[1] http://movielens.org

## 4.2  Μεθοδολογία και διαδικασία πειραμάτων

Προκειμένου να αξιολογήσουμε την απόδοση του LLFR στη σύσταση top-N αντικειμένων στους χρήστες, πραγματοποιούμε μια σειρά από πειράματα χρησιμοποιώντας το `Yahoo!Music` σύνολο δεδομένων.

Συγκρίνουμε τον LLFR απέναντι στον PureSVD και σε ακόμα τέσσερις διάσημους graph-based top-N αλγόριθμους συστάσεων: τον *average Commute Time* (CT) [18], τον *Pseudo-Inverse of the user-item graph Laplacian* (L†) [19], τον *Matrix Forest Algorithm* (MFA) [11], και τον *ItemRank* (IR) [23].

Ακολουθεί μία σύντομη περιγραφή για τον καθένα από τους παραπάνω.

**Average Commute Time (CT).**  Ως average commute time $n(i,j)$ ορίζεται ο μέσος αριθμός βημάτων τα οποία θα χρειαστεί ένας τυχαίος περιηγητής, ξεκινώντας από την κατάσταση $i \neq j$ για να φτάσει στην κατάσταση $j$ για πρώτη φορά και πίσω στην $i$. Ο *average Commute Time* (CT) χρησιμοποιεί το average commute time $n(i,j)$ για να ταξινομήσει τα στοιχεία του συνόλου που εξετάζεται, όπου $i, j$ είναι στοιχεία της βάσης δεδομένων. Για παράδειγμα, εάν πρόκειται για σύσταση ταινιών σε ανθρώπους, τότε ο αλγόριθμος υπολογίζει το average commute time ανάμεσα σε *ανθρώπους* στοιχεία και *ταινίες* στοιχεία. Όσο πιο μικρό είναι το αποτέλεσμα τόσο πιο όμοια είναι τα δύο στοιχεία [18]. Αν μετρήσουμε την απόσταση ανάμεσα στους κόμβους που αναπαριστούν ανθρώπους και ταινίες στο δοθέντα διμερή γράφο, μπορούμε να χρησιμοποιήσουμε αυτό το σκορ για την ταξινόμηση των ταινιών [23].

**Pseudo-Inverse of the user-item graph Laplacian (L†).**  Το μητρώο αυτό περιέχει τα εσωτερικά γινόμενα των διανυσμάτων που αντιστοιχούν στους κόμβους σε έναν Ευκλείδειο χώρο όπου οι κόμβοι είναι διαχωρισμένοι με βάση το commute time [18, 19].

**Matrix Forest Algorithm (MFA).**  Το MFA μητρώο περιέχει στοιχεία τα οποία παρέχουν επιπλέον μέτρα ομοιότητας ανάμεσα στους κόμβους του γράφου ενσωματώνοντας μη κατευθυνόμενα μονοπάτια, με βάση το matrix-forest θεώρημα [11].

**ItemRank (IR).**  Πρόκειται για μία μέθοδο συστάσεων που βασίζεται στον αλγόριθμο PageRank, η οποία παράγει ένα εξατομικευμένο διάνυσμα βαθμολογιών για το σύνολο των αντικειμένων, χρησιμοποιώντας έναν τυχαίο περίπατο με επανεκκινήσεις σε ένα γράφο συσχετίσεων μεταξύ αντικειμένων. Πιο συγκεκριμένα, ταξινομεί τις προτιμήσεις ενός χρήστη $u$ για νέα αντικείμενα $i$ ως την πιθανότητα ο $u$ να επισκεφτεί το $i$ κατά τη διάρκεια ενός τυχαίου περιπάτου σε ένα γράφημα του οποίου οι κόμβοι αντιστοιχούν στα αντικείμενα του συστήματος και οι ακμές συνδέουν αντικείμενα τα οποία έχουν βαθμολογηθεί από κοινούς χρήστες [17]. Η ιδέα στην οποία στηρίζεται ο αλγόριθμος ItemRank είναι ότι μπορεί να χρησιμοποιηθεί το μοντέλο που προκύπτει από το Γράφο Συσχετίσεων (Correlation Graph) για την πρόβλεψη των προτιμήσεων του χρήστη [23].

Στην περίπτωση των latent factor μεθόδων, δοκιμάσαμε τους αλγόριθμους για κάθε σύνολο δεδομένων χρησιμοποιώντας 20-800 latent factors και αναφέρουμε τα καλύτερα αποτελέσματα που επιτεύχθηκαν.

Για να αξιολογήσουμε την ποιότητα των συστάσεων, υιοθετήσαμε τη μεθοδολογία που προτάθηκε από τους συγγραφείς του [13]. Πιο συγκεκριμένα, οι βαθμολογίες χωρίζονται σε δύο υποσύνολα: στο σύνολο εκπαίδευσης $\mathcal{M}$ και στο σύνολο ελέγχου $\mathcal{T}$. Το $\mathcal{T}$ περιλαμβάνει μόνο βαθμολογίες με 5 αστέρια (άριστα), το οποίο μπορούμε με ασφάλεια να ισχυριστούμε ότι περιέχει αντικείμενα σχετικά με τους αντίστοιχους χρήστες. Στην περίπτωση που εξετάζουμε, το σύνολο εκπαίδευσης $\mathcal{M}$ είναι το αρχικό πλήρες σύνολο δεδομένων.

Συγκεκριμένα, πρώτα συλλέγουμε τυχαία το 1.4% των βαθμολογιών στο σύνολο δεδομένων προκειμένου να δημιουργήσουμε το σύνολο αξιολόγησης $\mathcal{P}$. Στη συνέχεια, χρησιμοποιούμε κάθε αντικείμενο $v_j$, το οποίο έχει βαθμολογηθεί με 5 αστέρια από το χρήστη $u_i$ στο $\mathcal{P}$, για να σχηματίσουμε το σύνολο ελέγχου $\mathcal{T}$. Τέλος, για κάθε αντικείμενο στο $\mathcal{T}$ που έχει βαθμολογηθεί με 5 αστέρια, επιλέγουμε τυχαία άλλα 1000 μη βαθμολογημένα αντικείμενα του ίδιου χρήστη (υποθέτουμε ότι πρόκειται για αντικείμενα που δεν ενδιαφέρουν άμεσα το χρήστη) και σχηματίζουμε λίστες κατάταξης ταξινομώντας και τα 1001 αντικείμενα σύμφωνα με τα σκορ συστάσεων που παράγονται από κάθε μέθοδο.

Έστω $p$ η θέση κατάταξης στη λίστα του αντικειμένου $v_j$ που εξετάζεται κάθε φορά. Το βέλτιστο αποτέλεσμα αντιστοιχεί στην περίπτωση όπου το $v_j$ βρίσκεται πιο πάνω στην κατάταξη από όλα τα τυχαία αντικείμενα ($p = 1$). Σχηματίζουμε την top-N λίστα συστάσεων επιλέγοντας τα πρώτα N αντικείμενα από τη λίστα (βρίσκονται πιο ψηλά στην κατάταξη). Αν $p \leq N$ έχουμε επιτυχία (hit), δηλαδή το αντικείμενο συστάθηκε στο χρήστη. Διαφορετικά, έχουμε αποτυχία (miss). Οι πιθανότητες επιτυχίας αυξάνονται όσο μεγαλώνει το N. Όταν $N = 1001$ έχουμε πάντα επιτυχία.

## 4.3 Μετρικές απόδοσης

Σε πολλές εφαρμογές το ΣΣ δεν προβλέπει τις προτιμήσεις του χρήστη για κάποια αντικείμενα, όπως οι βαθμολογίες ταινιών, αλλά προσπαθεί να συστήσει στους χρήστες αντικείμενα τα οποία είναι πιθανό να τους φανούν χρήσιμα. Για παράδειγμα, όταν ο χρήστης επιλέγει κάποιες ταινίες, το Netflix του προτείνει και άλλες ταινίες που μπορεί να τον ενδιαφέρουν με βάση αυτές που επέλεξε. Σε αυτήν την περίπτωση, δεν μας ενδιαφέρει αν το σύστημα πρόβλεψε σωστά ή όχι τις βαθμολογίες για αυτές τις ταινίες αλλά εάν το σύστημα πρόβλεψε σωστά ότι αυτές οι ταινίες θα ενδιαφέρουν τον χρήστη ώστε να τις επιλέξει.

Έτσι και στο top-N πρόβλημα. Δεν μας ενδιαφέρει η πρόβλεψη της βαθμολογίας που θα έδινε ο χρήστης σε ένα αντικείμενο, αλλά ενδιαφερόμαστε στο να παρουσιάσουμε στο χρήστη τα N αντικείμενα που θα ήθελε να χρησιμοποιήσει περισσότερο. Έτσι, επιλέξαμε και τις κατάλληλες μετρικές για την αξιολόγηση των αποτελεσμάτων μας, οι οποίες εστιάζουν στη μέτρηση της χρησιμότητας μιας ταξινομημένης λίστας αντικειμένων που παράγει το σύστημα συστάσεων στο χρήστη.

Για να ελέγξουμε λοιπόν την ποιότητα των συστάσεων, μεταξύ άλλων, χρησιμοποιούμε και τις καθιερωμένες μετρικές ακρίβειας **Recall** και **Precision**, όπως αυτές ορίστηκαν στο [13]. Όπως αναφέραμε και παραπάνω, για κάθε μία περίπτωση που εξετάζουμε (αποτελείται από 1001 αντικείμενα), έχουμε ένα και μοναδικό σχετικό αντικείμενο (το αντικείμενο εκείνο που ανήκει στο $\mathcal{T}$). Έτσι, η recall για μία συγκεκριμένη περίπτωση μπορεί να υποθέσει είτε την τιμή 0 (στην περίπτωση αποτυχίας) είτε την τιμή 1 (στην περίπτωση επιτυχίας). Ομοίως, η precision μπορεί να υποθέσει είτε την τιμή 0 είτε $1/N$. Τα συνολικά recall και precision ορίζονται λοιπόν από το μέσο όρο όλων των περιπτώσεων:

$$\text{recall(N)} = \frac{\#\text{hits}}{|\mathcal{T}|}$$

$$\text{precision(N)} = \frac{\#\text{hits}}{N \cdot |\mathcal{T}|}$$

όπου $|\mathcal{T}|$ το πλήθος των βαθμολογιών ελέγχου.

Αξίζει να σημειώσουμε σε αυτό το σημείο, ότι έχουμε υποθέσει πως όλα τα 1000 τυχαία αντικείμενα είναι άσχετα στο χρήστη και κατά συνέπεια, αυτή η υπόθεση οδηγεί στην υποβάθμιση των τιμών recall και precision που υπολογίζουμε σε σχέση με τις πραγματικές τιμές.

Για την αξιολόγηση των αποτελεσμάτων μας, χρησιμοποιούμε επίσης και άλλους γνωστούς δείκτες κατάταξης με βάση τη χρησιμότητα, οι οποίοι υποβαθμίζουν τη χρησιμότητα ενός αντικειμένου που συστήνεται κατά ένα παράγοντα που εξαρτάται από τη θέση του στη λίστα συστάσεων [49]. Με άλλα λόγια, τα αντικείμενα τα οποία δε βρίσκονται ψηλά στη λίστα τιμωρούνται πιο αυστηρά καθώς είναι πιο πιθανό ο χρήστης να μην τα παρατηρήσει.

Η χρησιμότητα κάθε σύστασης είναι η χρησιμότητα του αντικειμένου που συστήνεται μειωμένη κατά ένα παράγοντα που εξαρτάται από τη θέση του στη λίστα με τις συστάσεις. Ένα παράδειγμα τέτοιας χρησιμότητας είναι η πιθανότητα ο χρήστης να παρατηρήσει μία σύσταση στη θέση $i$ της λίστας. Υποτίθεται συνήθως ότι οι χρήστες παρατηρούν τις λίστες συστάσεων από την αρχή ως το τέλος, με τη χρησιμότητα των συστάσεων να φθίνει πιο έντονα καθώς κινούμαστε προς το τέλος της λίστας. Αυτή η μείωση μπορεί επίσης να ερμηνευτεί ως η πιθανότητα ότι ένας χρήστης θα παρατηρούσε μία σύσταση σε μία συγκεκριμένη θέση στη λίστα, με τη χρησιμότητα της σύστασης, δεδομένου ότι παρατηρήθηκε, να εξαρτάται μόνο από το αντικείμενο που συστήθηκε. Υπό αυτήν την έννοια, η πιθανότητα ότι μία συγκεκριμένη θέση στη λίστα συστάσεων παρατηρήθηκε, υποτίθεται ότι εξαρτάται μόνο από τη θέση και όχι από τα αντικείμενα που συστήνονται [49].

Σε πολλές εφαρμογές, ο χρήστης μπορεί να χρησιμοποιήσει είτε μόνο ένα είτε ένα μικρό αριθμό αντικειμένων από αυτά που συστήνονται. Σε τέτοιες περιπτώσεις, αναμένεται οι χρήστες να παρατηρήσουν μόνο λίγα αντικείμενα τα οποία βρίσκονται στην κορυφή της λίστας συστάσεων. Τέτοιες εφαρμογές μπορούν να μοντελοποιηθούν χρησιμοποιώντας σημαντική μείωση σε σχέση με τη θέση καθώς κινούμαστε προς τα κάτω στη λίστα. Μία υποψήφια μετρική για αυτές τις περιπτώσεις είναι το **R-Score**.

Σε άλλες εφαρμογές ο χρήστης αναμένεται να διαβάσει ένα σχετικά μεγάλο μέρος της λίστας. Σε κάποια είδη αναζήτησης, όπως η αναζήτηση νομικών εγγράφων, οι χρήστες ίσως να ψάχνουν για όλα τα σχετικά αντικείμενα, και να ήταν πρόθυμοι να διαβάσουν μεγάλο μέρος της λίστας συστάσεων [49].

Αυτές οι μετρικές παρουσιάζονται παρακάτω:

Έστω $\pi_q$ ο δείκτης του $q^{th}$ αντικειμένου στη λίστα κατάταξης με συστάσεις $\boldsymbol{\pi}$, και $\mathbf{y}$ ένα διάνυσμα με τιμές σχετικότητας για μια ακολουθία αντικειμένων.

- **R-Score.** Η μετρική R-Score υποθέτει ότι η αξία των συστάσεων φθίνει εκθετικά στη λίστα κατάταξης και δίνει για κάθε χρήστη την ακόλουθη βαθμολογία:

$$\text{R-Score}(\alpha) = \sum_q \frac{\max(y_{\pi_q} - d, 0)}{2^{\frac{q-1}{\alpha-1}}}$$

  όπου $d$ είναι μία "δεν ενδιαφέρομαι" βαθμολογία που εξαρτάται από το πρόβλημα, και $\alpha$ είναι μία half-life παράμετρος, η οποία ελέγχει την εκθετική πτώση της αξίας των θέσεων στη λίστα κατάταξης [49].

- **NDCG@$k$.** Το Cumulative Discounted Gain αποτελεί ένα μέτρο κατά το οποίο οι θέσεις κατάταξης φθίνουν λογαριθμικά και ορίζεται ως:

$$\text{DCG@}k(\mathbf{y}, \boldsymbol{\pi}) = \sum_{q=1}^{k} \frac{2^{y_{\pi_q}} - 1}{\log_2(2 + q)}$$

  Το κανονικοποιημένο Discounted Cumulative Gain μπορεί συνεπώς να οριστεί ως ο λόγος του DCG@$k(\mathbf{y}, \boldsymbol{\pi})$ ως προς το DCG της καλύτερης δυνατής κατάταξης (βλ. [3] για λεπτομέρειες).

- **Mean Reciprocal Rank.** MRR είναι ο μέσος όρος του reciprocal rank score κάθε χρήστη, και ορίζεται ως ακολούθως:

$$\text{RR} = \frac{1}{\min_q\{q : y_{\pi_q} > 0\}}$$

## 4.4 Ποιότητα Συστάσεων

Ακολουθώντας τη διαδικασία ελέγχου των Karypis et al. [25], παίρνουμε τα `Yahoo!Music` δεδομένα και δημιουργούμε δύο σύνολα δεδομένων με διαφορετική πυκνότητα (η πυκνότητα προκύπτει από το πλήθος των βαθμολογιών δια το γινόμενο του πλήθους χρηστών και αντικειμένων). Αυτό συνέβη ώστε να αξιολογηθεί συγκεκριμένα η απόδοση των αλγορίθμων σε δεδομένα χαμηλής πυκνότητας, τα οποία συναντώνται αρκετά συχνά σε ρεαλιστικά σενάρια.

Πιο συγκεκριμένα, επιλέγουμε τυχαία ένα υποσύνολο από χρήστες και αντικείμενα από τα συνολικά. Στη συνέχεια, διατηρώντας το ίδιο πλήθος χρηστών και αντικειμένων, δημιουργούμε το πρώτο πιο αραιό σύνολο δεδομένων διαγράφοντας τυχαία τιμές από το μητρώο με τις βαθμολογίες. Όμοια, δημιουργείται και το δεύτερο ακόμα πιο αραιό σύνολο δεδομένων. Συγκεκριμένα, αφαιρούμε με τυχαίο τρόπο τιμές από το μητρώο βαθμολογιών του αραιότερου συνόλου δεδομένων. Τα σύνολα δεδομένων που προκύπτουν σημειώνονται ως `Yahoo1` και `Yahoo2` και οι πυκνότητές τους είναι $1.63\%$ και $0.55\%$ αντίστοιχα.

Στην πρώτη γραμμή του Σχήματος 4.1, αναφέρουμε το Recall ως συνάρτηση του $N$ (δηλαδή τον αριθμό των αντικειμένων που συστήνονται) και στη δεύτερη γραμμή το Precision ως συνάρτηση του Recall, για το `Yahoo1` (πρώτη στήλη) και `Yahoo2` (δεύτερη στήλη). Σε ό,τι αφορά το Recall($N$), έχουμε εστιάσει σε τιμές του $N$ στο εύρος $[1 \ldots 20]$. Μεγαλύτερες τιμές του $N$ μπορούν να αγνοηθούν για ένα τυπικό top-N πρόβλημα [13]. Στην τελευταία γραμμή του σχήματος, αναφέρουμε το Normalized Discounted Cumulative Gain για top-N λίστες κατάταξης, και πάλι για τιμές του $N$ στο εύρος $[1 \ldots 20]$.

Όπως μπορούμε να δούμε, ο LLFR έχει πολύ καλή απόδοση φτάνοντας τη δεύτερη θέση στο πυκνό `Yahoo1` και την πρώτη θέση στο `Yahoo2` σύνολο δεδομένων. Πιο συγκεκριμένα, βλέπουμε ότι ο LLFR καταφέρνει να διατηρήσει την απόδοσή του στην πιο αραιή περίπτωση φτάνοντας για παράδειγμα, για N = 15, τιμή Recall περίπου 0.32, ενώ ο PureSVD πέφτει στο 0.25. Αυτό πρακτικά σημαίνει ότι περίπου 32% των 5-άστερων αντικειμένων παρουσιάζονται από τον LLFR στις top-15 λίστες κατάταξης για τους αντίστοιχους χρήστες, ένα αποτέλεσμα ικανοποιητικό δεδομένης της αραιότητας των δεδομένων. Τα ίδια αποτελέσματα ισχύουν και για τις μετρικές Precision και NDCG. Επίσης, αξίζει να σημειωθεί ότι οι μέθοδοι τυχαίων περιπάτων, ItemRank και Commute Time, τα πάνε ιδιαίτερα καλά, ειδικά στα πιο αραιά δεδομένα όπου έχουν καλύτερη απόδοση από τον PureSVD σε όλες τις μετρικές.

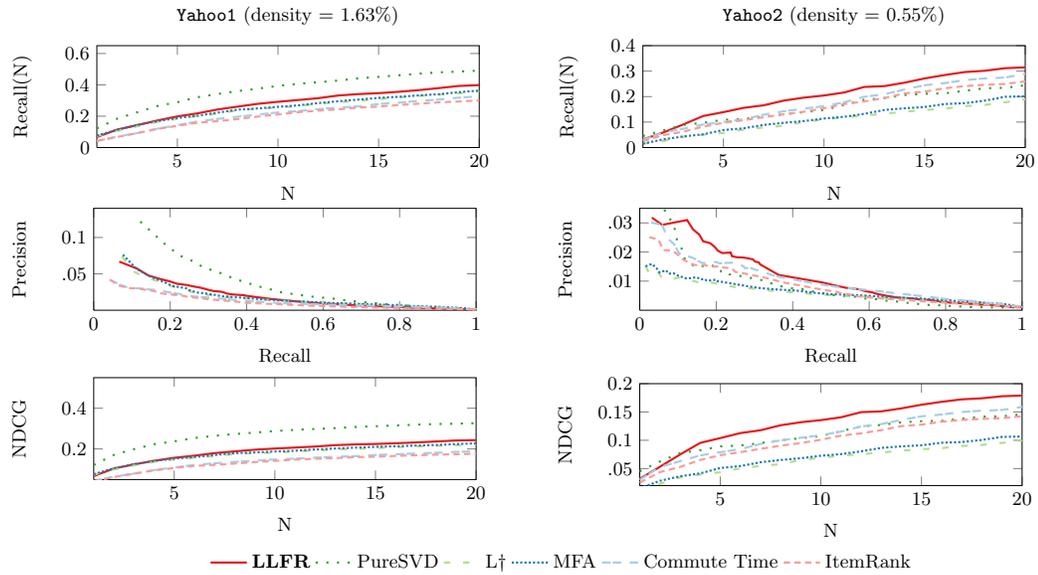

ΣΧΉΜΑ 4.1: Αξιολόγηση της απόδοσης top-N συστάσεων.

Παρατηρούμε ότι ο LLFR παράγει πολύ καλά αποτελέσματα, και ταυτόχρονα είναι κατά πολύ ο πιο συμφέρων υπολογιστικά σε σχέση με όλες τις μεθόδους που εξετάστηκαν. Να σημειώσουμε σε αυτό το σημείο ότι ο αλγόριθμος Commute Time απαιτείται να χειριστεί ένα γράφημα με $n+m$ ακμές (όπου $n$ είναι ο αριθμός των χρηστών και $m$ ο αριθμός των αντικειμένων) και να υπολογίσει $2nm$ βαθμολογίες κατά το πρώτο πέρασμα. Ομοίως, οι L† και MFA, απαιτούν τον άμεσο υπολογισμό του ψευδο-αντίστροφου του Laplacian, ενός γραφήματος με $n+m$ ακμές, και την αντιστροφή ενός $(n+m)$-διάστατου τετραγωνικού μητρώου, αντίστοιχα (βλέπε [18] για περισσότερες λεπτομέρειες). Πρόκειται για προβλήματα τα οποία γίνονται εύκολα μη διαχειρίσιμα καθώς ο αριθμός των χρηστών στο σύστημα μεγαλώνει. Πράγματι, μόνο οι LLFR και ItemRank περιλαμβάνουν μητρώα των οποίων οι διαστάσεις εξαρτώνται μόνο από τη διάσταση του χώρου αντικειμένων, η οποία στις περισσότερες πραγματικές εφαρμογές αυξάνει αργά. Ωστόσο, στα πειράματά μας, παρατηρούμε ότι ο LLFR τρέχει 10 φορές πιο γρήγορα από τον ItemRank[2]. Αυτό ισχύει για κάθε σύνολο δεδομένων που δοκιμάσαμε.

Τέλος, το γεγονός ότι η μέθοδός μας ξεπερνά σε απόδοση όλες τις άλλες μεθόδους στο πιο αραιό σύνολο δεδομένων υποδεικνύει ότι ο LLFR θα μπορούσε να είναι πολύ χρήσιμος για την αντιμετώπιση ενός πολύ βασικού προβλήματος, το οποίο είναι γνωστό ότι ευθύνεται για τη σημαντική υποβάθμιση της ποιότητας σε πραγματικά συστήματα συστάσεων, το *Cold Start Problem*.

---

[2]επιλέγοντας convergence tolerance $10^{-5}$, damping factor 0.85 (όπως προτείνεται από τους συγγραφείς στο [23]), και υπολογίζοντας τα διανύσματα συστάσεων για όλους τους χρήστες με τη μία, ώστε να εκμεταλλευτούμε τα βελτιστοποιημένα BLAS3 kernels.

## 4.5 Το Πρόβλημα της Κρύας Εκκίνησης

Το cold start πρόβλημα έχει να κάνει με τη δυσκολία πραγματοποίησης αξιόπιστων συστάσεων εξαιτίας μιας αρχικής έλλειψης βαθμολογιών [8]. Πρόκειται για ένα πρόβλημα που συναντάται συχνά σε πραγματικά συστήματα συστάσεων κατά τα πρώτα τους στάδια, όπου δεν υπάρχουν αρκετές βαθμολογίες προκειμένου οι αλγόριθμοι συνεργατικής διήθησης να ανακαλύψουν ομοιότητες ανάμεσα σε αντικείμενα και χρήστες (*New Community Problem*).

Ωστόσο, το πρόβλημα μπορεί να προκύψει και κατά την εισαγωγή νέων χρηστών σε ένα ήδη υπάρχον σύστημα (*New Users Problem*). Όπως είναι αναμενόμενο, επειδή αυτοί οι χρήστες είναι νέοι, δεν έχουν βαθμολογήσει ακόμα πολλά αντικείμενα και έτσι, ο CF αλγόριθμος δεν μπορεί ακόμα να κάνει αξιόπιστες προσωποποιημένες συστάσεις. Κάτι τέτοιο μπορεί να θεωρηθεί ως ένα είδος προβλήματος τοπικής αραιότητας και αποτελεί μία από τις διαρκείς προκλήσεις που αντιμετωπίζουν τα συστήματα συστάσεων σε λειτουργία [28].

Τέλος, συχνά παρουσιάζεται και το πρόβλημα λόγω νέων αντικειμένων (*New Items Problem*) [8], όπου τα νέα αντικείμενα που μπαίνουν στο σύστημα συνήθως δεν έχουν αρχικά βαθμολογίες, και έτσι δεν είναι πιθανό να συσταθούν σε κάποιο χρήστη. Κατά συνέπεια, ένα αντικείμενο το οποίο δεν περιλαμβάνεται σε συστάσεις, περνάει απαρατήρητο από μεγάλο μέρος της κοινότητας των χρηστών, και καθώς οι χρήστες δε γνωρίζουν την ύπαρξή του, δεν το βαθμολογούν. Με αυτόν τον τρόπο, προκύπτει ένας φαύλος κύκλος, όπου αντικείμενα του συστήματος παραμερίζονται και δεν αποτελούν μέρος της διαδικασίας συστάσεων. Μία κοινή λύση σε αυτό το πρόβλημα είναι να δίνεται κίνητρο σε κάποιους χρήστες ώστε να βαθμολογούν κάθε νέο αντικείμενο που εισέρχεται στο σύστημα [8].

Εμείς στην παρούσα εργασία ασχολούμαστε με τις δύο πρώτες εκδοχές του προβλήματος. Η τρίτη εκδοχή του προβλήματος όπως αυτή αναφέρθηκε παραπάνω (καθώς και το πλήρες πρόβλημα), μελετάται στο [44], όπου προσομοιώνεται η κατάσταση και εξετάζεται η απόδοση του προτεινόμενου αλγορίθμου HIR.

### 4.5.1 New Community Problem

Προκειμένου να αξιολογήσουμε την απόδοση του LLFR στην αντιμετώπιση του new community problem, εκτελούμε το ακόλουθο πείραμα: προσομοιώνουμε το φαινόμενο επιλέγοντας τυχαία να συμπεριλάβουμε το 10%, 20% και 30% του `Yahoo1` συνόλου δεδομένων σε τρεις νέες τεχνητά αραιές εκδόσεις, με τέτοιο τρόπο όπου κάθε σύνολο δεδομένων είναι υποσύνολο του επόμενου. Η ιδέα είναι ότι αυτά τα νέα σύνολα δεδομένων αναπαριστούν στιγμιότυπα των αρχικών σταδίων του συστήματος συστάσεων, όταν η κοινότητα των χρηστών ήταν νέα και το σύστημα υστερούσε σε βαθμολογίες.

Στη συνέχεια, παίρνουμε τα νέα σύνολα δεδομένων και δημιουργούμε σύνολα ελέγχου ακολουθώντας τη μεθοδολογία που περιγράφεται στο Κεφάλαιο 4.2. Εκτελούμε όλους τους αλγόριθμους και αξιολογούμε την απόδοσή τους χρησιμοποιώντας το Mean Reciprocal Rank και το R-Score με halflife $\alpha = 5$. Επιλέξαμε αυτές τις μετρικές λόγω του γεγονότος ότι μπορούν να συνοψίσουν την απόδοση συστάσεων σε έναν μόνο αριθμό, κάτι που κάνει ευκολότερη τη σύγκριση της top-N ποιότητας για τα διαφορετικά στάδια κατά την εξέλιξη του συστήματος. Ο Πίνακας 4.1 παρουσιάζει τα αποτελέσματα.

ΠΙΝΑΚΑΣ 4.1: Αποτελέσματα απόδοσης για το *New Community* Πρόβλημα.

|  | **LLFR** | PureSVD | L† | MFA | CT | IR |
|---|---|---|---|---|---|---|
| *10%* | | | | | | |
| MRR | **0.1184** | 0.1075 | 0.0106 | 0.0571 | 0.0197 | 0.0870 |
| R-Score | **0.1474** | 0.1296 | 0.0085 | 0.0563 | 0.0089 | 0.1028 |
| *20%* | | | | | | |
| MRR | **0.0874** | 0.0722 | 0.0257 | 0.0271 | 0.0459 | 0.0630 |
| R-Score | **0.1238** | 0.1180 | 0.0309 | 0.0331 | 0.0728 | 0.0905 |
| *30%* | | | | | | |
| MRR | **0.0930** | 0.0924 | 0.0316 | 0.0348 | 0.0646 | 0.0741 |
| R-Score | **0.1352** | 0.1289 | 0.0396 | 0.0454 | 0.1047 | 0.1117 |

Βλέπουμε ξεκάθαρα ότι ο LLFR έχει καλύτερη απόδοση από κάθε άλλο αλγόριθμο και στα τρία στάδια, με το προβάδισμά του να είναι μεγαλύτερο στις δύο πιο αραιές περιπτώσεις. Να σημειωθεί ότι το σύνολο δεδομένων που χρησιμοποιήθηκε είναι το Yahoo1, ένα σύνολο δεδομένων στο οποίο ο PureSVD "είχε το πάνω χέρι", καθώς όπως αποδείχτηκε είχε καλύτερη απόδοση στην πλήρη περίπτωση που παρουσιάστηκε στην προηγούμενη ενότητα.

### 4.5.2 New Users Problem

Για να αξιολογήσουμε την απόδοση του αλγορίθμου μας στην αντιμετώπιση του new users problem, χρησιμοποιούμε ξανά το Yahoo1 σύνολο δεδομένων και εκτελούμε το ακόλουθο πείραμα. Επιλέγουμε τυχαία 50 χρήστες οι οποίοι έχουν βαθμολογήσει 100 αντικείμενα ή περισσότερα, και διαγράφουμε τυχαία το 95% των βαθμολογιών τους. Το σκεπτικό είναι ότι τα τροποποιημένα δεδομένα αναπαριστούν μία "προηγούμενη έκδοση" του συνόλου δεδομένων, όταν οι

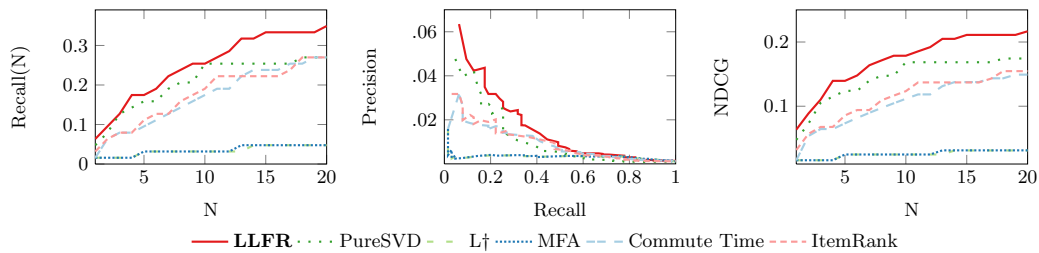

ΣΧΗΜΑ 4.2: Αξιολόγηση της απόδοσης top-N συστάσεων για το *New Users* Πρόβλημα

χρήστες ήταν νέοι στο σύστημα και έτσι, είχαν λιγότερες βαθμολογίες. Στη συνέχεια, παίρνουμε το μέρος του συνόλου δεδομένων που αντιστοιχεί σε αυτούς τους νέους χρήστες και δημιουργούμε το σύνολο ελέγχου όπως και προηγουμένως, χρησιμοποιώντας 10% ως cut-off αυτή τη φορά για το σύνολο αξιολόγησης, ώστε να έχουμε αρκετές ταινίες βαθμολογημένες με 5 στο σύνολο ελέγχου και να εκτιμήσουμε αξιόπιστα την ποιότητα της απόδοσης. Μία παρόμοια μέθοδος χρησιμοποιήθηκε στο [37] για την αξιολόγηση της Web ranking ποιότητας για το πρόβλημα των νέων σελίδων που προστίθενται (*Newly Added Pages Problem*), όπου οι σελίδες αυτές είναι νέες και έτσι έχουν λιγότερα εισερχόμενα links και κατά συνέπεια, το link graph είναι αραιό.

Στο Σχήμα 4.2, βλέπουμε ότι ο LLFR έχει τις καλύτερες επιδόσεις συγκριτικά με όλους τους άλλους αλγορίθμους που εξετάστηκαν σε όλες τις μετρικές. Αξίζει να σημειωθεί, ότι παρότι ο PureSVD έχει καλή απόδοση σε αυτό το σύνολο δεδομένων, και η πυκνότητα της τροποποιημένης έκδοσης παραμένει κοντά στην αρχική (1.46%), η ποιότητά του για το σύνολο των νέων χρηστών έχει μειωθεί σημαντικά. Επιπλέον, είναι ενδιαφέρον το ότι ο LLFR καταφέρνει να έχει σημαντικά καλύτερη απόδοση από όλες τις graph-based μεθόδους, οι οποίες θεωρούνται στη βιβλιογραφία ανάμεσα στις πιο υποσχόμενες προσεγγίσεις για την αντιμετώπιση των προβλημάτων που σχετίζονται με την αραιότητα, λόγω της ικανότητάς τους να εκμεταλλεύονται τις μεταβατικές σχέσεις στα δεδομένα [17, 23].

## Κεφάλαιο 5

# Συμπεράσματα

Στην παρούσα διπλωματική εργασία παρουσιάσαμε τον αλγόριθμο *Lanczos Latent Factor Recommender*. Πρόκειται για μία νέα εναλλακτική latent factor-based πρόταση για το πρόβλημα των top-N συστάσεων, υπολογιστικά αποδοτική, και κατάλληλη για εφαρμογές μεγάλου όγκου δεδομένων (*big data*). Το βασικό χαρακτηριστικό του LLFR το οποίο του δίνει αυτό το προβάδισμα κόστους, είναι ότι μειώνει τη διάσταση του προβλήματος κατασκευάζοντας τη *Lanczos βάση* του Krylov υποχώρου που ορίζεται από ένα μητρώο συσχετίσεων μεταξύ αντικειμένων. Στη συνέχεια, χρησιμοποιεί το χαμηλής διάστασης μοντέλο για να παράγει λίστες κατάταξης αντικειμένων για τον κάθε χρήστη.

Πραγματοποιήσαμε μία σειρά από πειράματα σε πραγματικά σύνολα δεδομένων, και συγκεκριμένα στα `MovieLens10M` και `Yahoo!Music`, και συγκρίναμε τον LLFR με άλλους διάσημους αλγόριθμους, οι οποίοι φημίζονται για την ικανότητά τους να τα πηγαίνουν καλά στις διάφορες προκλήσεις που χαρακτηρίζουν τις σύγχρονες εφαρμογές. Για την αξιολόγηση των αποτελεσμάτων επιλέξαμε ευρέως χρησιμοποιούμενες μετρικές, οι οποίες έχουν νόημα για ένα πραγματικό σύστημα συστάσεων.

Τα πειράματα έδειξαν ότι ο LLFR επιτυγχάνει πολύ καλά αποτελέσματα απέναντι στις άλλες μεθόδους συνεργατικής διήθησης με τις οποίες συγκρίθηκε, σε επίπεδο τόσο υπολογιστικού κόστους όσο και ποιότητας συστάσεων. Αξίζει δε να τονίσουμε ότι η μέθοδός μας συμπεριφέρεται ιδιαίτερα καλά όταν η αραιότητα των δεδομένων είναι έντονη, όταν δηλαδή δεν υπάρχουν αρκετά δεδομένα στο σύστημα ώστε ο αλγόριθμος να ανακαλύψει συσχετίσεις μεταξύ χρηστών και αντικειμένων, και να πραγματοποιήσει επιτυχημένες συστάσεις, όπως στο Πρόβλημα Κρύας Εκκίνησης (*Cold Start Problem*). Η αραιότητα είναι ένα σύνηθες πρόβλημα στις σύγχρονες πραγματικές εφαρμογές διότι συνήθως οι χρήστες αλληλεπιδρούν μόνο με ένα μικρό ποσοστό των διαθέσιμων αντικειμένων, και την ίδια στιγμή νέοι χρήστες και νέα αντικείμενα προστίθενται τακτικά στο σύστημα.

Για να ελέγξουμε την απόδοση του αλγόριθμού μας στο πρόβλημα Κρύας Εκκίνησης, πραγματοποιήσαμε πειράματα προσομοιώνοντας την περίπτωση *New Community Problem* – η οποία συναντάται στα πραγματικά συστήματα κατά τα αρχικά τους στάδια όπου δεν υπάρχουν αρκετά δεδομένα ακόμα στο σύστημα – και *New Users Problem* – η οποία συναντάται κατά την εισαγωγή νέων χρηστών σε ένα υπάρχον σύστημα, όπου ακριβώς επειδή αυτοί οι χρήστες είναι νέοι δεν έχουν προλάβει να βαθμολογήσουν αντικείμενα. Τα αποτελέσματα έδειξαν ότι ο LLFR τα πάει καλύτερα σε σχέση με όλες τις άλλες μεθόδους που εξετάστηκαν, συμπεριλαμβανομένων και των graph-based τεχνικών οι οποίες είναι πολλά υποσχόμενες στην αντιμετώπιση της αραιότητας.

Η απλοϊκότητα του LLFR φαίνεται να είναι πιο αποδοτική στην περίπτωση αραιών δεδομένων, σε σχέση με άλλους πιο ακριβείς ή έξυπνους ή πολύπλοκους αλγόριθμους, όπως ο PureSVD, οι οποίοι πολλές φορές αποτυγχάνουν στην απόδοσή τους λόγω του overfitting που παρουσιάζουν σε συνθήκες αραιότητας.

Συνοψίζοντας, τα αποτελέσματά μας υποδεικνύουν ότι τόσο το υπολογιστικό προφίλ του LLFR όσο και η απόδοσή του στο top-N πρόβλημα συστάσεων τον καθιστούν ως έναν πολύ καλό υποψήφιο για ευρείας κλίμακας εφαρμογές συστάσεων.

# Βιβλιογραφία